\documentclass[preprint,12pt]{elsarticle}
\usepackage{amssymb}
\usepackage{amsmath}
\usepackage{placeins}
\usepackage{longtable}
\usepackage{adjustbox}

\journal{ }

\begin{document}

\begin{frontmatter}

\title{Augmented Structure Preserving Neural Networks for
cell biomechanics}

\author[label1]{Juan Olalla-Pombo} 
\author[label1]{Alberto Badías} 
\author[label1]{Miguel Ángel Sanz-Gómez} 
\author[label1]{José María Benítez} 
\author[label1,label2]{Francisco Javier Montáns}

\affiliation[label1]{organization={E.T.S. de Ingeniería Aeronáutica y del Espacio},
            addressline={Universidad Politécnica de Madrid, \\Plaza del Cardenal Cisneros, 3},
            city={Madrid},
            postcode={28040},
            country={Spain}}

\affiliation[label2]{organization={Department of Mechanical and Aerospace Engineering, Herbert Wertheim College of Engineering},
            addressline={University of Florida},
            city={Gainesville},
            postcode={FL 32611},
            country={USA}}

\begin{abstract}
Cell biomechanics involve a great number of complex phenomena that are fundamental to the evolution of life itself and other associated processes, ranging from the very early stages of embryo-genesis to the maintenance of damaged structures or the growth of tumors. Given the importance of such phenomena, increasing research has been dedicated to their understanding, but the many interactions between them and their influence on the decisions of cells as a collective network or cluster remain unclear. We present a new approach that combines Structure Preserving Neural Networks, which study cell movements as a purely mechanical system, with other Machine Learning tools (Artificial Neural Networks), which allow taking into consideration environmental factors that can be directly deduced from an experiment with Computer Vision techniques. This new model, tested on simulated and real cell migration cases, predicts complete cell trajectories following a roll-out policy with a high level of accuracy. This work also includes a mitosis event prediction model based on Neural Networks architectures which makes use of the same observed features. 
\end{abstract}


\end{frontmatter}

\section{Introduction}
\label{sec: introduction}

Cell migration mechanisms are known to be present in many fundamental processes throughout the evolution of living organisms. From the creation of life through embryogenesis \cite{horwitz2003cell, kurosaka2008cell} to the growth and reparation of tissue structures or the evolution of tumors \cite{friedl2009collective, torres2018histatins, polacheck2013tumor}, the movement of cells is necessary, and the collective patterns followed by cell networks and clusters determine not only the kinetics but also the results of these processes.

As cells are living units that perform complex tasks, undergo constant reactions and transformations, interact with other cells and can respond to their surroundings, their migration can be presented as the result of a large number of internal and external factors. The influence of many environmental factors such as chemical gradients that can be created with biomaterials \cite{wu2012gradient} or that might appear in organic environments \cite{dormann2003chemotactic}, density gradients caused by cell accumulation \cite{iglesias2008navigating}, or even the presence of dead cells (which can be of interest in wound healing or tumor growth processes) \cite{hu1970theory} has been thoroughly studied. Other external factors related to cell collective movement and the tensile forces that can appear between them have also been analyzed \cite{trepat2011plithotaxis, gjorevski2015dynamic}, with several works in this field showing that cells can use protuberances to attach themselves to other cells, which later exert pulling or pushing forces to guide their movement \cite{AMIEVA1995180}. 

Despite the precision that the proposed models can achieve while explaining the relation between these factors and cell movements, there is a general lack of a global approach to the problem. Due to the possible interrelations between environmental properties, many studies simplify the problem by creating conditions where the studied gradient or feature is the dominant source of instability, and thus the main reason behind cell migration \cite{saucedo2024simple}. This often requires the design of \textit{in vitro} experiments that take place in synthetic environments that aim to resemble organic conditions \cite{benetti2016mimicking}. However, the environmental conditions inside living beings or surrounding single-cell organisms are complex and encompass a large number of gradients (nutrients, oxygen, or other types of chemical variation) \cite{malijauskaite2021gradients}, which cause cells to follow different trends of movement. The number of \textit{in vivo} experiments is scarce, due to their execution complexity and the rapid development of \textit{in vitro} techniques \cite{keenan2008biomolecular}, which impedes a proper understanding of the complete underlying mechanisms of cell migration. In these cases, it can be difficult to correctly assess the magnitude of the gradients or possible interactions between cells, resulting in high levels of uncertainty in the observations.

To overcome these issues, we present a new approach that aims to understand cell migration mechanisms as a whole, making use of Machine Learning (ML) techniques. ML algorithms allow pattern detection in large sets of data and can handle interrelated features that influence one another. Thus, their application to complex and fluctuating problems (such as organic environments) is becoming more common. Extensive efforts have been made in the fields of tissue engineering, where these methods can help predict cell behavior and development \cite{rafieyan2023mlate}. Some works have even applied Deep Learning (DL) networks to cell migration problems. García-Moreno \textit{ et al.} \cite{garcia2024using} propose  recurrent and convolutional network architectures to study the migration mechanisms of breast cancer cells in the context of wound healing. Their work has been tested on \textit{in vivo} cases that showcase the ability of this type of techniques to reproduce complex trajectories. Other studies have also used ML techniques to adjust known physical phenomena that participate in cell migration processes (such as nematic mechanisms or hydrodynamic effects \cite{colen2021machine}). Even if these results have been validated and are a promising tool to predict tissue evolution and cell movement, they do not provide as much information on the underlying physical mechanisms due to the opacity of recurrent units. These models used a supervised learning approach that focuses on the minimization of a loss function, without taking into account the compliance with any physical principle. Whereas forcing the compliance of experimental data with adjusted laws of kinematics and dynamics can be a reduced approach that oversees a great number of physical and biological effects that influence cells, biomechanical systems must still comply with universal principles such as the laws of thermodynamics.

In this context, we propose a hybrid model that combines Structure Preserving Neural Networks (SPNN) with Multi Layer Perceptrons (MLP), named Augmented Structure Preserving Neural Networks (ASPNN). Its objective is to predict cell velocities for a given moment in time and entire trajectories following a \textit{roll-out} policy (meaning that the position of the cell is only provided to the model as data in the first frame) given a series of inputs related to cell state and surrounding conditions. The SPNN studies the problem from a purely mechanical standpoint, and allows compliance of the system with thermodynamical principles: conservation of total energy and increasing entropy. This submodel focuses on the deep analysis of the energetic behavior of the system, allowing the detection of external sources of energy, such as chemical gradients of some kind. However, as mentioned before, cell migration problems include a series of factors that extend beyond the mechanical domain and are related to cellular responses to external cues. Many of said factors can be controlled or observed in \textit{in vitro} experiments, such as some chemical or density gradients \cite{mak2011microfabricated}, but are not always defined \textit{in vivo}. However, some cannot be continuously observed, such as the degree of attachment between cells (some studies measure it qualitatively by stress maps on a substrate \cite{schlie2012dynamics} or visual analysis of shape and area variations \cite{webb2000relationships}). Cell systems are thus not completely observable, which forces the application of a different tool that allows underlying pattern detections. Our MLP aims to learn the influence of external factors beyond purely mechanical events (such as elastic collisions) on the migration behavior of cells. The low level of comprehensibility of the learning process of this architecture is aligned with the inherent complexity of the biological processes that take place. 

With this approach, we combine explainable studies of a physical system that complies with basic principles with a detailed analysis of a large number of factors simultaneously, eliminating the need for \textit{ad hoc} crafted environments that highlight a given gradient or external stimuli. Moreover, the model presented only uses features that can be obtained from a video or a set of images as input. These features are extracted by means of Computer Vision (CV) techniques for image segmentation and object tracking. That allows its application to \textit{in vivo} environments and experiments, since there is no need for additional data. Hence, we propose a general method that allows noninvasive (visual observations can serve as input) \textit{in vivo} analysis of cell migration from a global point of view, with a high degree of explainability. This model has been tested on synthetic cases and a real \textit{in vitro} experiment. Furthermore, we present a mitotic event prediction model to determine the probability of a cell to undergo a mitotic process given its current state and environmental cues (using the same features as the ASPNN). In this regard, some studies have focused on the detection, analysis, and recognition of mitotic events using DL with high precision results (see the work by Mao \textit{et al.} \cite{mao2019cell} or Delgado-Rodríguez \textit{ et al.} \cite{delgado2024automatic}), but very few have explored the predictive task. Having the ability to correctly predict division events would allow better understanding and forecasting of the evolution of rapidly growing systems, such as tumor environments, where the mitotic rate can be a defining parameter \cite{szabo2013cellular}.

The data acquisition and processing process is described in Section \ref{sec:methods}, while Section \ref{sec:results} contains a discussion of the results, and final comments are included in Section \ref{sec:conclusion}.

\section{Materials and Methods}
\label{sec:methods}

This section includes a general overview of the process for data acquisition and the creation of the neural model architecture. 

\subsection{Data processing and feature extraction}

The input for our ASPNN is directly a video sequence, that has to be
divided consecutive frames. A Segment Anything Model (SAM) \cite{kirillov2023segany} architecture was used to segment the image, locate and create a mask for each cell on the frame. Image segmentation was the preferred alternative over other methods of cell location, such as object detection because the extracted information about the shape and size of the cell was deemed of interest to characterize its behavior (e.g., cells usually undergo shape variations when starting mitosis processes). Frame images first underwent a series of preprocessing modifications, such as the application of median blur filters to eliminate noise \cite{opencv_library}, Wiener deconvolution filters to increase resolution \cite{van2014scikit}, or grayscale. Table \ref{tab:data} shows the data obtained from segmentation.

\begin{longtable}{|p{0.4\linewidth}|p{0.4\linewidth}|}
\caption{Data to be obtained for each cell and frame in the segmentation stage.}
\label{tab:data}\\
\hline
\textbf{Parameter}         & \textbf{Comments}                           \\ \hline
\endfirsthead
\multicolumn{2}{c}%
{{\bfseries Table \thetable\ continued from previous page}} \\
\endhead
\textbf{Cell ID Number}    & Automatically assigned                      \\ \hline
\textbf{Segmentation Mask} & Stored in Run-Length Enconding (RLE) format \\ \hline
\textbf{Cell Area}         & Number of pixels covered by the mask        \\ \hline
\textbf{Eccentricity}      & Cells are approximated to elliptical bodies \\ \hline
\textbf{Predicted Intersection over Union (IoU)} & To be used as predicted accuracy to discard incorrect segmentations \\ \hline
\textbf{Stability Score}   & Confidence score value                      \\ \hline
\end{longtable}

\begin{figure}[htb]
    \centering
    \includegraphics[width=0.9\linewidth]{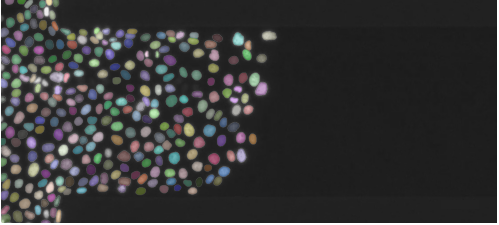}
    \caption{Segmented image with masks represented as colored layers over cells in the frame.}
    \label{fig:sam}
\end{figure}

After optimization of a number of parameters (mainly predicted Intersection over Union threshold and stability score threshold), SAM models showed great results in cell segmentation, recognizing cells with very low levels of brightness or separating clusters of several of them with diffused cell borders (see Figure \ref{fig:sam}).

As a requirement of the tracking algorithms, the bounding boxes were also calculated for each cell mask, using the maximum and minimum horizontal and vertical coordinates to determine its limits. In this case, IoU threshold values between bounding boxes were included so that cells in clusters could be efficiently detected and separated. During the first tests, additional bounding boxes were created around the entire clusters. In addition, only cells located within the vertical limits of the microchannel and at a $50$ pixel distance from the left border of the image were considered, to avoid partially visible cells or those that enter and exit the frame, which could distort the function of the tracking algorithms (they were included in images, but data from those bounding boxes were discarded).

\FloatBarrier

Once cells have been detected and segmented in each of the frames, their trajectories must be tracked to gain understanding of their movement. This poses a complex problem due to the high number of cells in the channel, their \textit{a priori} high variability movement, and the similarities between them, which makes differentiation tasks (based on size and shape) difficult. After several preliminary tests, the selected tracking algorithm was DeepOCSort \cite{maggiolino2023deep} with ReIdentification (ReID) models. This algorithm, based on OC-Sort \cite{cao2022observation}, uses Kalman Filters and ReIdentification models based on visual features to associate objects in subsequent frames. The model had to be finely tuned to correctly track cells along the sequence. General Intersection over Union (GIoU) was the preferred association function, with threshold levels of $0.05$ for the addition of a mask to a given trajectory. Recovery parameters were also included; for example, if a given trajectory ends due to the lack of suitable candidates in a frame, candidates will still be searched for in the following $4$ frames. This was found to be useful for correctly tracking objects in the limits of the microchannel. The general tendency was to allow low-confidence detection of a large number of trajectories, which have later been filtered on the basis of stability, length, and visual correlation with the image sequence. Table \ref{tab:datatracking} shows data obtained from the tracking algorithm.

\begin{longtable}{|p{0.4\linewidth}|p{0.4\linewidth}|}
\caption{Data to be obtained for each cell and frame in the tracking stage.}
\label{tab:datatracking}\\
\hline
\textbf{Parameter}           & \textbf{Comments}                                \\ \hline
\endfirsthead
\multicolumn{2}{c}%
{{\bfseries Table \thetable\ continued from previous page}} \\
\endhead
\textbf{Bounding Box Limits} & Coordinates of top-left and bottom-right corners \\ \hline
\textbf{ID Number}           & For trajectory identification                    \\ \hline
\textbf{Confidence}          & For trajectory filtering                         \\ \hline
\textbf{Frame Number}        & Later used for data classification               \\ \hline
\end{longtable}

After filtering the trajectories (eliminating those with fewer than $5$ frames), a total of $566$ different trajectories were obtained. Other filters were applied in subsequent stages, since some new trajectories might have been created due to mitotic events or loss of a given cell during a series of frames. Trajectories data were then grouped by trajectory and by frame, which was useful to later determine environmental conditions (such as number of surrounding cells) for a cell in a given frame.

After data were collected for cell trajectories, a number of features were extracted directly from cell positions and shape parameters (area and eccentricity). These features will then be used as input to the ASPNN. Given the structure of the model, only trajectories with more than $105$ frames were considered correct trajectories (for data encapsulation matters, only the first $105$ frames were studied for these trajectories). Data from the rest of the cells were still used to calculate environmental characteristics. For every frame of each of the correct trajectories, Table \ref{tab:datafeatures} shows the obtained features.

\begin{longtable}{|p{0.4\linewidth}|p{0.4\linewidth}|}
\caption{Features obtained for each frame of the correct trajectories.}
\label{tab:datafeatures}\\
\hline
\textbf{Feature} & \textbf{Comments} \\ \hline
\endfirsthead
\multicolumn{2}{c}%
{{\bfseries Table \thetable\ continued from previous page}} \\
\endhead
\textbf{Cell center coordinates} & x and y coordinates \\ \hline
\textbf{Cell velocity} & In x and y directions, calculated as the difference between the center coordinates of the cell in subsequent frames. \\ \hline
\textbf{Density gradient} & Calculated from the collective area of cells located in a $8$ 20-pixel square grid surrounding the cell. \\ \hline
\textbf{Cell density in each grid square} & $8$ values corresponding to the density of cells on each region. \\ \hline
\textbf{Number of surrounding cells} & Calculated as the number of cells with center coordinates within 75 pixels of the cell. \\ \hline
\textbf{Number of cells per sector} & The images have been divided into $4$ different sectors (top left, top right, bottom left, and bottom right). The number of cells in each sector provides data on the general distribution of cells and free space. \\ \hline
\textbf{Average velocity of surrounding cells} & In $x$ and $y$ directions. \\ \hline
\textbf{Cell brightness value} & Obtained directly from the grayscale images. \\ \hline
\textbf{Cell area variation} & Difference in cell area in subsequent frames. \\ \hline
\textbf{Cell eccentricity} &  \\ \hline
\end{longtable}

These parameters would later be divided into output (cell velocity) and input (all remaining features), resulting in vectors of size $23$ and $2$, respectively, for each frame of each correct trajectory.

\subsection{ASPNN model}
Our Augmented Structure Preserving Neural Network consists of two parts: a structure preserving model based on the General Equation for the Nonequilibrium Reversible-Irreversible Coupling, commonly referred to as GENERIC \cite{ottinger1997dynamics}, and an MLP model, named CoNN (Correction Neural Network), considering environmental conditions surrounding the cell.

\subsubsection{GENERIC submodel}
GENERIC algorithm defines a problem structure based on two "potentials", total energy ($E$) and entropy ($S$). With this definition, the mechanical problem can be divided into reversible and irreversible dynamics. Reversible dynamics are contemplated within the context of total energy conservation (energy term), whereas irreversible dynamics is governed by the imposed increase in entropy. Each of the potentials is linked to a matrix that contains the relations between the observable variables of the system. For a system with a set of observable variables \textbf{$z$}, the evolution of said variables over time can be calculated with

\begin{equation}
    \label{eq:generic basic}
    \frac{d\textbf{z}}{dt} = L\frac{\partial E}{\partial \textbf{z}}+M\frac{\partial S}{\partial \textbf{z}} \ ,
\end{equation}

where $L$ and $M$ are the linear operators related to total energy and entropy, respectively, and which must comply with the following degeneracy conditions to ensure thermodynamical consistency:

\begin{equation}
    L\frac{\partial S}{\partial \textbf{z}} = 0 \ ,
\end{equation}
\begin{equation}
    M\frac{\partial E}{\partial \textbf{z}} = 0 \ .
\end{equation}

The structure of both matrices will depend on the structure of the problem, but they can usually be derived from the literature or, for conservative systems, can be constructed from resulting equations of Hamiltonian mechanics applied to the system. However, these matrices might be unknown for complex non-observable systems with irreversible terms.

Within the theoretical framework of GENERIC, many different numerical methods for mechanical analysis have emerged during the past years. In this context, the work by Romero \cite{romero2009thermodynamically} is of great interest. The author proposes time-stepping algorithms that allow propagation of the state of a system over time, which leads to the characterization of its mechanical behavior. These integrating algorithms allow for the transformation of the problem into a set of discrete time steps by means of finite differences, so that Equation \ref{eq:generic basic} translates to

\begin{equation}
    \label{eq:generic operators}
    \frac{\textbf{z}_{n+1}-\textbf{z}_{n}}{\Delta t} = L\cdot \frac{\text{D} E}{\partial \textbf{z}} + M\cdot \frac{\text{D} S}{\partial \textbf{z}} \ ,
\end{equation}

where $\text{D}E$ and $\text{D}S$ are the discrete gradients of both potentials.

Over the past few years, many efforts have been guided towards the combination of the GENERIC framework and Deep Learning. While providing general rules for the evolution of mechanical systems, they can be adapted to complex systems where the exact governing mechanisms are not exactly known, as they must still be thermodynamically consistent. The authors of this manuscript have studied the possible applications of this combination in previous work \cite{hernandez2021structure}. Their major contribution is the new definition of discrete energy and entropy gradients in terms of a gradient matrix, based on a finite element approach,

\begin{equation}
    \frac{\text{D} E}{\partial \textbf{z}} \simeq A\textbf{z} \ ,
\end{equation}
\begin{equation}
    \frac{\text{D} S}{\partial \textbf{z}} \simeq B\textbf{z} \ ,
\end{equation}

so that the resulting equation for the evolution of the system would be

\begin{equation}
    \label{eq:generic final}
    \frac{\textbf{z}_{n+1}-\textbf{z}_{n}}{\Delta t} = L\cdot A\textbf{z} + M\cdot B\textbf{z}
\end{equation}

and degeneracy conditions result in

\begin{equation}
    L\cdot B\textbf{z} = 0
\end{equation}

\begin{equation}
    M\cdot A\textbf{z} = 0
\end{equation}

where $A$ and $B$ are gradient matrices of which the structure is determined to adapt to the structure of the problem during the learning process. By learning these two matrices, the model is learning the underlying physics of the problem (relations between observable variables and energy and entropy variations) while ensuring compliance with thermodynamical laws. The model structures proposed in this study are based on those presented in the cited work.

Thus, the first part of the ASPNN consists of a MLP which will output the corresponding $A$ and $B$ matrices. For the training process, a loss function that encompasses both the correct prediction of state variables and the correct evolution of thermodynamical potentials in following time-steps has been defined. This loss function ($l_{t}$) is calculated by a linear combination of data loss ($l_{data}$), which will compare the prediction of the entire ASPNN with the ground truth by

\begin{equation}
    l_{data} = (\textbf{z}_{pred}-\textbf{z}_{gt})^{2}
\end{equation}

where $\textbf{z}_{pred}$ and $\textbf{z}_{gt}$ are predicted and ground truth values for state variables, respectively; and degeneracy loss ($l_{deg}$), calculated as

\begin{equation}
    l_{deg} = ||L\cdot B_{pred}\textbf{z}_{pred}||_{2}^{2} + ||M\cdot A_{pred}\textbf{z}_{pred}||_{2}^{2}
\end{equation}

which results in

\begin{equation}
    l_{t} = \lambda_{d}l_{data} + l_{deg} \ ,
\end{equation}

where $\lambda_{d}$ is a tunable parameter that will be adjusted depending on the magnitude of each of the loss terms.

Our SPNN takes the position (size 2 vector with coordinates $x$ and $y$) and the velocities ($\dot{x}$ and $\dot{y}$), which results in a vector of input variables of size 4. For this set of variables, $L$ matrix, which must be skew-symmetric, is defined \cite{romero2009thermodynamically}:

\begin{equation}
    L = \begin{bmatrix}
0 & 0 & 1 & 0\\
0 & 0 & 0 & 1 \\
-1 & 0 & 0 & 0 \\
0 & -1 & 0 & 0 
\end{bmatrix}	
\end{equation}

As for the matrix $M$, which is required to be symmetric and positive semi-definite (to ensure positive entropy variations), it has been defined as

\begin{equation}
    M = \begin{bmatrix}
1 & -0.5 & 0 & 0\\
-0.5 & 1 & 0 & 0 \\
0 & 0 & 1 & -0.5 \\
0 & 0 & -0.5 & 1 
\end{bmatrix}
\end{equation}

to account for possible dissipation related to cell position and velocity (friction due to viscous effects) in both coordinates (which can be intertwined, thus the $-0.5$ terms outside of the main diagonal).

This SPNN GENERIC-based architecture will output a new set of variables for each time-step (position and velocity in the following frame). The model is based on a fully connected architecture with $5$ hidden layers with $16, 64, 128, 64$ and $16$ neurons. Every layer uses hyperbolic tangent activation functions for smooth derivability. In order to maximize its range, inputs have been normalized to a $[-1,1]$ interval 

\begin{equation}
    \label{eq:normalization}
    x_{norm} = 2\cdot \frac{x-\max(x)}{\max(x)-\min(x)} - 1 \ .
\end{equation}

\subsubsection{CoNN submodel}
\begin{figure}[htb]
    \centering
    \includegraphics[width=\linewidth]{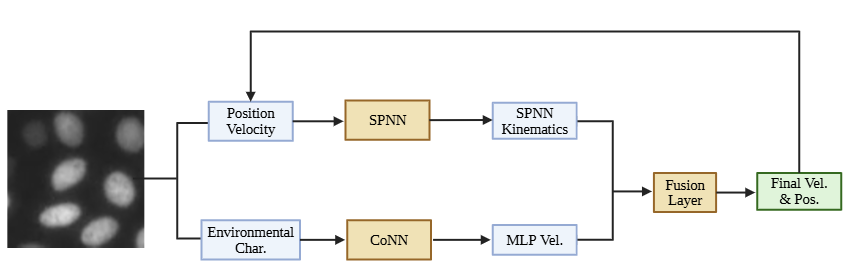}
    \caption{Complete process pipeline. Note that position and velocities inputs to the SPNN are only provided from the video in the first frame, with results from the model being used in subsequent frames.}
    \label{fig: pipeline}
\end{figure}

The resulting velocities will then be combined with the aforementioned MLP model considering environmental characteristics. The final cell velocity will be obtained from the combination (with a multineuron layer) of both velocities. This velocity will then be used to calculate the new position of the cell (which will then be used as input in the next frame), following a \textit{roll-out} policy. The complete process pipeline is shown in Figure \ref{fig: pipeline}.

This part of the model encompasses a fundamental part of the complexity of the underlying physical phenomena involved in cell migration, which is a process that surpasses the purely mechanical realm and is guided and influenced by surrounding conditions and gradients. This combination of responses to the external environment by the cell does not fully follow the physical properties of simple systems due to the large number of actors and internal mechanisms taking place. Thus, in order to fully understand the overall behavior of the system, an additional element with the ability to learn without the thermodynamical constraints of the GENERIC model. Hence, this part of the model is responsible of absorbing the effect of the surroundings of the cell in its movement.

The complexity of the surroundings, which translates into high-variance input data, guides the size and importance of this part of the model, which has been set as a MLP due to the flexibility and simplicity of training of this kind of architecture. This allows robust learning processes that can be thoroughly controlled and guided despite the complexity of input data and their unclear relations with output data (cell kinematics).

The MLP model consists of a fully connected architecture with four hidden layers with 181, 297, 149 and 295 neurons. The number of layers for this model, as well as the number of neurons for each one and the learning rate have been optimized with a Bayesian Optimization and HyperBand (BOHB) algorithm, which tests a series of parameter sets to determine which minimizes the defined loss function \cite{falkner2018bohb}.

The final MLP layer translates a set of $4$ inputs (velocities from SPNN and MLP) into $2$ final outputs. Although the weights and biases for this layer will still be updated during the learning process, its reduced size has made the strategy shift towards a conservative approach with low learning rates. All three parts of the model have been trained for $5000$ epochs. The learning rates and learning rate scheduler parameters have been fine-tuned for each case, and are presented in the corresponding epigraphs of Section \ref{sec:results}.  

\subsection{Mitosis prediction model}
The mitosis prediction model also uses a MLP architecture with fully connected layers. It uses the same input as the ASPNN, that is, the position and velocity of a cell (also obtained from model results in previous frames, following the \textit{roll-out} policy) and its surrounding conditions (environmental characteristics). In addition, two new inputs have been included, cell area and brightness variations over time. The reason for the introduction of these parameters was the visual analysis of the mitotic events in the video. Before undergoing a division, cells modify their shape, reducing their size and becoming more elliptical, and increasing their brightness. Cell shape variations during mitotic processes are characterized in the literature \cite{thery2006cell}. The theoretical reason for the relationship between cell brightness increases and the start of mitotic processes was not as clear. After a review of the literature, it was found that chromatin present in cell nuclei, which takes its structure due to attachment to H2B histones, undergoes restructurings during mitosis \cite{ito2007role, belmont1999large}. In the real \textit{in vitro} experiment used, the authors used mCherry-labeled H2B histones, which are present in the nuclei and therefore allow for clear identification of the nuclei. These histones have been extensively used as fluorescent markers \cite{giancotti1978intrinsic}. Thus, it is clear that the increase in brightness is related to the movements of chromatin in the cell, which can be closely related to the start of mitosis.

Thus, this model makes use of the same features (in addition to additional characteristics of the cell) obtained during the feature extraction process to predict mitotic events. This analysis is embedded in the framework described in the present work since these events are known to be related to the energetic cycles of the cells. In this case, a different part of the energetic state of the cell is used for cellular division instead of motility, which motivates the inclusion of this architecture in the present theoretical model.

The model has $4$ hidden layers with sizes $48, 96, 64$ and $32$, and as the chosen activation function has also been $\tanh$, the inputs have also been normalized to the range $[-1, 1]$ using Equation \ref{eq:normalization}. This model outputs two probabilities (ranging from $0$ to $1$), for a positive or negative mitotic event. Since it can be considered a classification problem, the activation function for the last layer has been Softmax, which is better suited for binary and multiclass classification tasks. This activation function tends to maximize the probability for one of the classes and minimize it for the rest of them.

Mitosis probabilities are predicted for every frame in the trajectory with no further restrictions, to force the model to learn the dynamics of these processes, which usually extend for more than $1$ frame. As frames with mitotic events have a value of $1$ for the ground truth and frames without mitosis have a value of $0$, the loss function has been calculated with a Binary Cross Entropy (BCE) loss, which is adequately suited for binary classification problems \cite{ruby2020binary}. Mitosis events have been labeled manually.

\section{Experiments and results}
\label{sec:results}
This section includes the results for the prediction of trajectories in the three cases (both artificial and real) and predictions of mitosis events in the real experiment.

\textit{In-silico} cases were created by simulating the movement of rigid circular two-dimensional bodies along a channel 300 pixels wide and 100 pixels high. This movement was stimulated in Python by an imposed velocity increase of 0.05 pixels per frame in every frame for the horizontal dimension and a 0.005 pixel per frame increase in the vertical dimension. A sequence of 100 frames was simulated. Conditions were applied to avoid overlapping of simulated cells, and all collisions were considered elastic. Data for the first case (cell position along the entire trajectory) were obtained numerically. A second Gaussian noise experiment was included in the extracted positions, and the resulting velocities and characteristics were calculated using these new positions. The aim of these artificial cases was to train the model on progressively complex and variant data.

Regarding the \textit{ in vitro} experiment, a video from Marel et al. \cite{marel2014flow} study on diffusive and density guided cell movement was used. Visual data from the experiment carried out by the authors to test their mathematical model was available in the additional data section. This video shows a time lapse (190 seconds) of Madine-Darby canine kidney cells (MDCK) moving along a two-dimensional channel. They used bright-field and fluorescence microscopy, which provided high visibility of cell nuclei and borders, which was considered a main factor in training the model on a clear dataset. The authors designed the experiment to force the two-dimensional movement of the cells to be in a predominant direction by creating narrow PEG-DMA microchannels. Cells were cultivated on a Petri dish surrounded by these channels, which were kept inaccessible. The channels were then opened, which triggered the movement of cells towards regions with lower density. In contrast with artificially created cases, cell position was not known as data beforehand. In order to obtain said data, a staged process of image segmentation for cell detection and object tracking for trajectory determination was proposed. Further processing of position and trajectory data was required to obtain environmental features that would be used as input to the MLP (also for artificial cases). Thus, the entire process could be divided into the following stages: image segmentation, cell tracking, data processing, ASPNN model, and mitosis prediction model. Only data processing and the ASPNN model apply to artificial cases.

\subsection{\textit{In-silico} determistic experiment}
The first artificial case is the simplest, since cell movement is only guided by a two-dimensional gradient without any noise. This deterministic experiment only includes an underlying equation for the movement of cells, thus being completely deterministic. By processing the data with the GENERIC architecture, the model is able to correctly infer this underlying mechanism for movement.

Since the trajectories are simple, there is a clear path towards loss minimization, which has encouraged the use of high learning rates that allow the SPNN architecture to reach a close prediction in a reduced number of epochs. Moreover, weights in the final MLP layer were initialized manually so that the SPNN architecture was taken as the dominant input. The learning rates for each of the parts of the model and their learning rate scheduler parameters are shown in Table \ref{tab: params case no noise}. As for MLP, its lesser importance has guided the use of lower learning rates and slower reductions.

\begin{longtable}{|c|c|c|c|}
\caption{Training parameters for the first artificial case. LR Scheduler epochs show the number of epochs between each update of the learning rate, which is carried out by multiplying the current learning rate by $\gamma$.}
\label{tab: params case no noise} \\
\hline
\centering
\textbf{Model} & \textbf{LR} & \textbf{LR Scheduler epochs} & \textbf{LR Scheduler $\gamma$} \\ \hline
\endhead
SPNN            & $1\cdot10^{-2}$ & $500$ & $0.1$ \\ \hline
MLP             & $1\cdot10^{-3}$ & $500$ & $0.2$ \\ \hline
Final MLP Layer & $1\cdot10^{-2}$ & $500$ & $0.1$ \\ \hline
\end{longtable}

\begin{figure}[htb]
    \centering
    \includegraphics[width=0.49\linewidth]{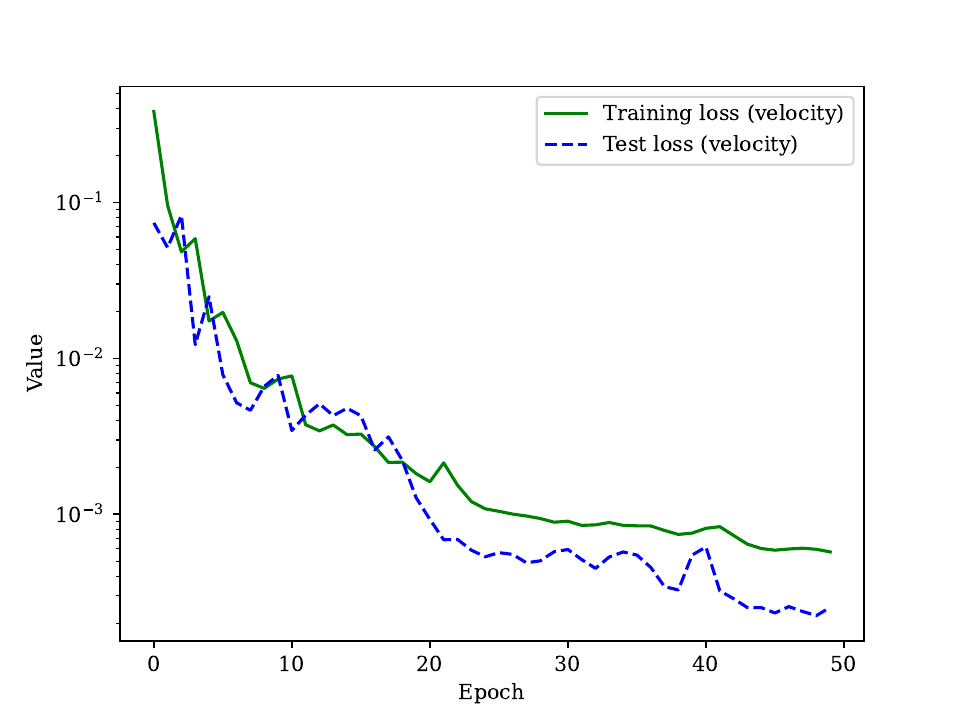}
    \includegraphics[width=0.49\linewidth]{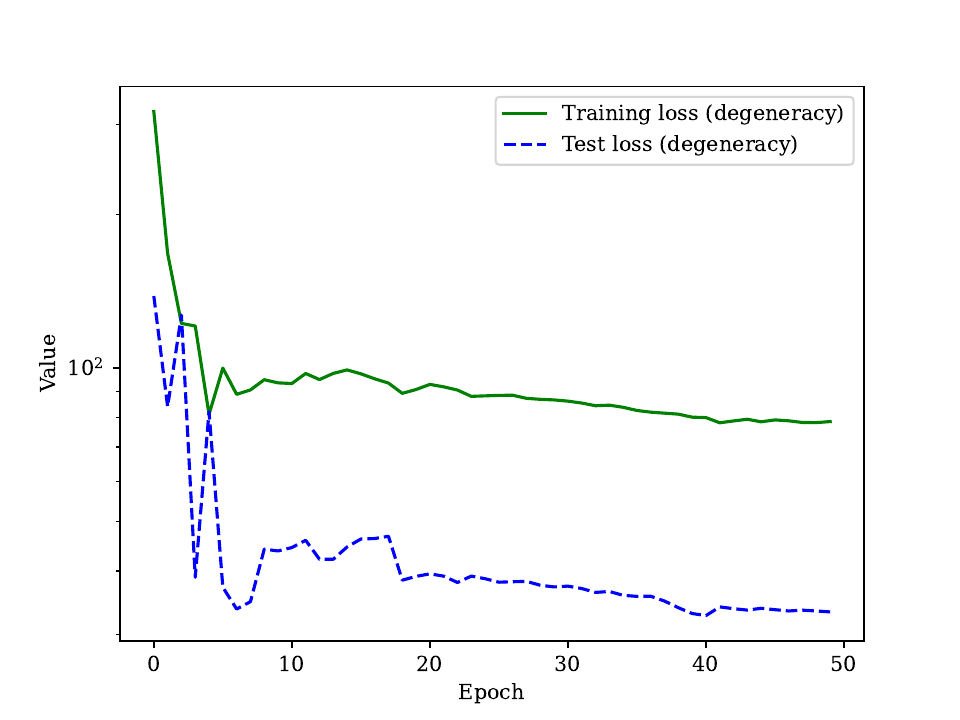}
    \caption{Evolution of loss functions for data (left) and degeneracy conditions (right) for the first artificial case. The horizontal axis represents hundreds of epochs.}
    \label{fig:loss case no noise}
\end{figure}

The learning process has been successful, with losses for both data and degeneracy conditions decreasing over epochs (see Figure \ref{fig:loss case no noise}). The model has shown high accuracy for the velocities of $90.9\%$ and $96.8\%$ for the coordinates $x$ and $y$. The lower value for accuracy in the $x$ coordinate could be caused by the higher gradient in this direction. This accuracy has been calculated by using the relative error,

\begin{equation}
    Acc = 100\cdot(1-|\frac{v_{pred}-v_{gt}}{v_{gt}}|) \ ,
\end{equation}

\begin{figure}[htb]
    \centering
    \includegraphics[width=0.49\linewidth]{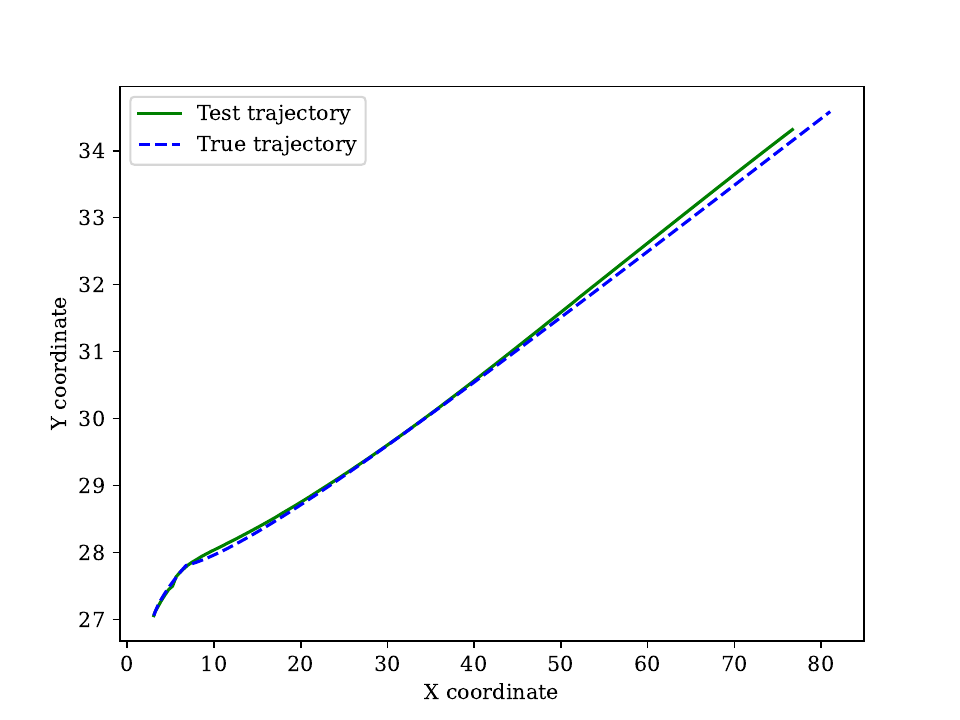}
    \includegraphics[width=0.49\linewidth]{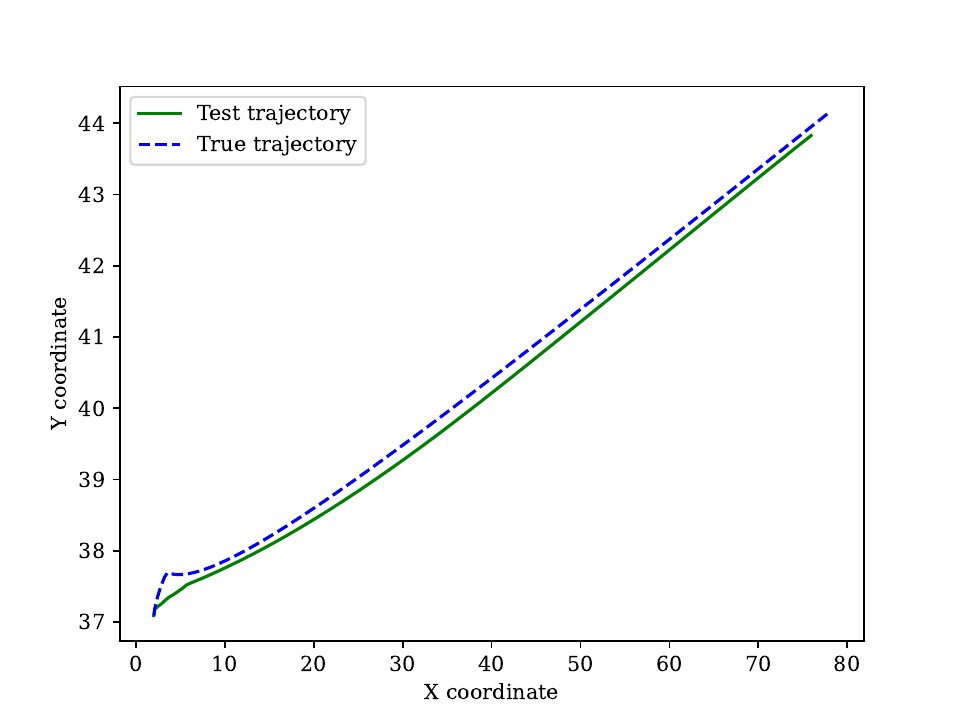}
    \caption{Comparison of two predicted and real trajectories for the first \textit{in-silico} case.}
    \label{fig: traj case no noise}
\end{figure}

where $v$ represents velocities. These high values are corroborated by the resulting trajectories, with two examples shown in Figure \ref{fig: traj case no noise}.

\begin{figure}[htb]
    \centering
    \includegraphics[width=0.49\linewidth]{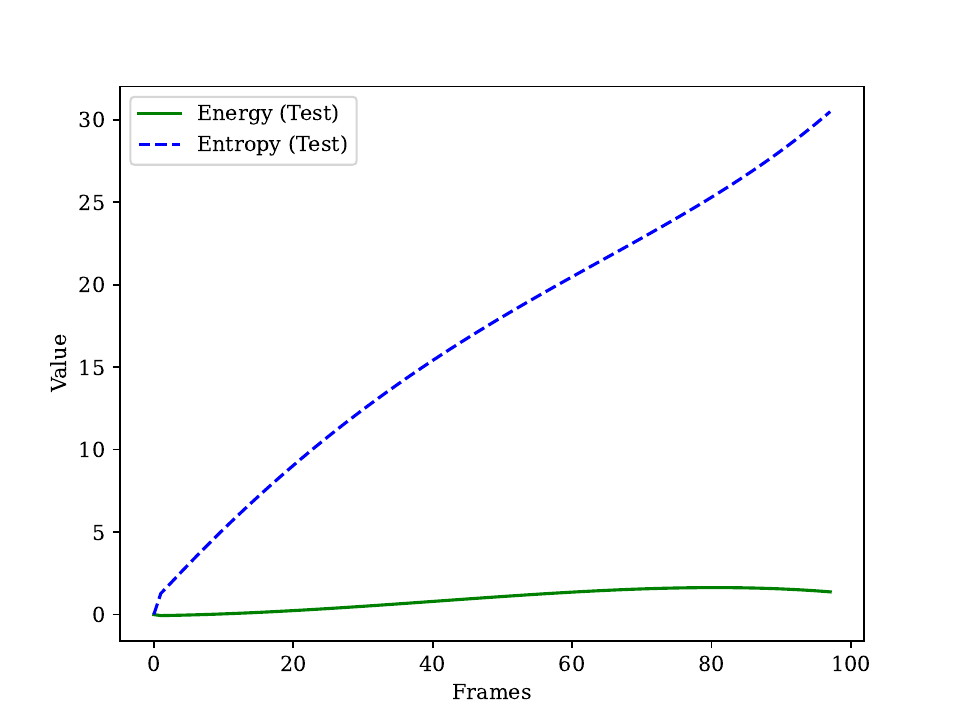}
    \caption{Evolution of thermodynamical variables for a complete trajectory.}
    \label{fig: thrmd case no noise}
\end{figure}

Moreover, the correct trajectory predictions have been combined with compliance with the thermodynamical properties, as shown in Figure \ref{fig: thrmd case no noise}. The total energy is not exactly constant but does not show an increasing tendency, while entropy only grows throughout the trajectory.

\begin{figure}[htb]
    \centering
    \includegraphics[width=0.49\linewidth]{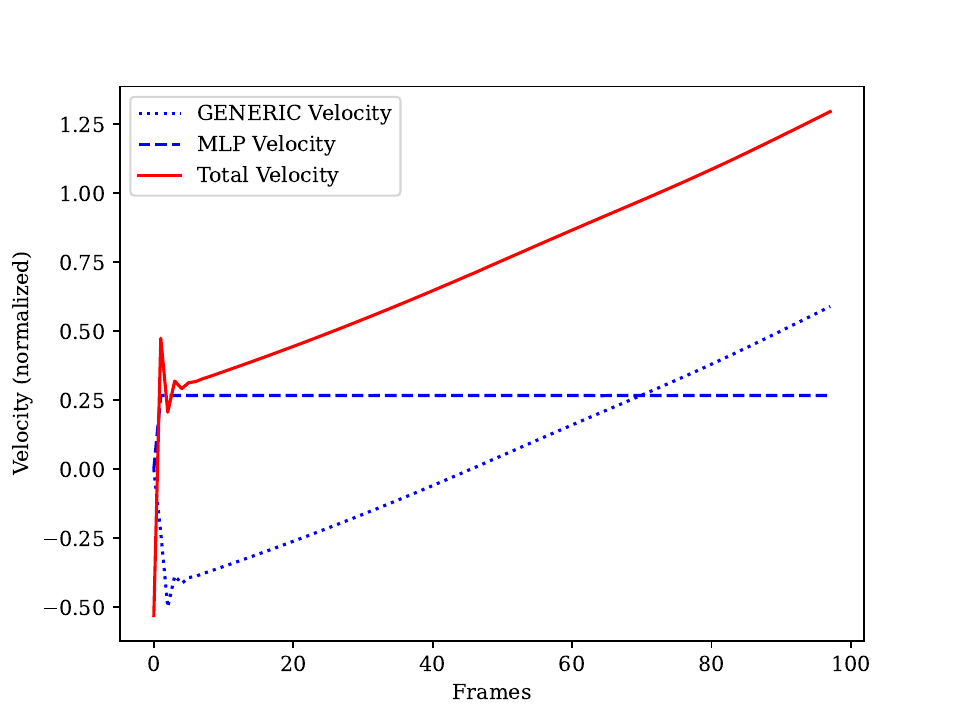}
    \includegraphics[width=0.49\linewidth]{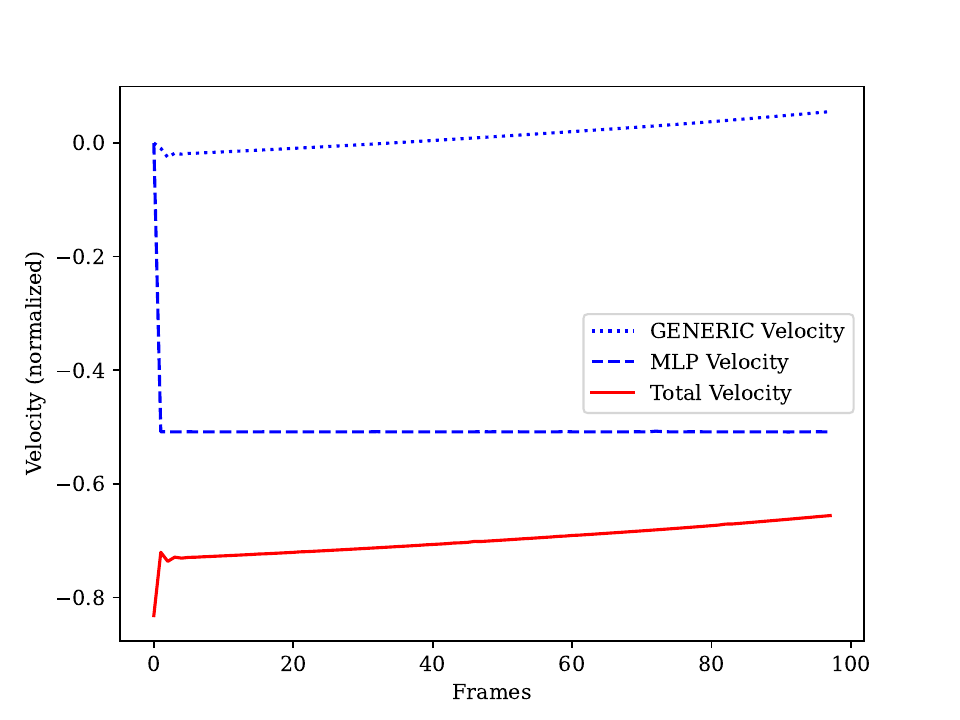}
    \caption{Contribution of each part of the model to the final output velocities for the \textit{in-silico} case without noise. SPNN shows an increasing tendency for both coordinates, while MLP provides variations.}
    \label{fig: vels case no noise}
\end{figure}

Since environmental conditions do not influence movement in this case, SPNN is the main output producer. Figure \ref{fig: vels case no noise} shows the contribution to the production of output of each part of the model throughout the trajectory, calculated as the product between the input value to the last neuron and its weights. It is noticeable that the SPNN (GENERIC) carries most of the weight, with the MLP providing slight variations. The SPNN captures the main trend of the velocity of the cells, with the MLP providing a constant offset, which could be understood as an initial velocity.

This first experiment proves that the proposed architecture is able to learn the underlying basic mechanisms for movement relying mostly in the GENERIC-based architecture, stating that the energetic behavior of these bodies is correctly depicted.

\FloatBarrier

\subsection{\textit{In-silico} case with noise}

The second case is very similar to the first one, only including random Gaussian noise to the data obtained from the simulation. The noise distribution had a maximum value of 10\% of the current velocity of the cell and was added before calculating new cell positions. As explained before, this case was created to assess the generalization capacity of the model, since real experiment data will be noisy given the accumulated errors and deviations in the image segmentation, tracking, and data processing stages. Thus, initial weights were reduced and learning strategies were more conservatively designed, since the SPNN architecture will not have a clear path towards solution optimization. Training parameters are shown in Table \ref{tab: params case noise}.

\begin{longtable}{|c|c|c|c|}
\caption{Training parameters for the second artificial case.}
\label{tab: params case noise} \\
\hline
\centering
\textbf{Model} & \textbf{LR} & \textbf{LR Scheduler epochs} & \textbf{LR Scheduler $\gamma$} \\ \hline
\endhead
SPNN            & $1\cdot10^{-2}$ & $500$ & $0.1$ \\ \hline
MLP             & $1\cdot10^{-2}$ & $500$ & $0.1$ \\ \hline
Final MLP Layer & $1\cdot10^{-2}$ & $500$ & $0.1$ \\ \hline
\end{longtable}

\begin{figure}[htb]
    \centering
    \includegraphics[width=0.49\linewidth]{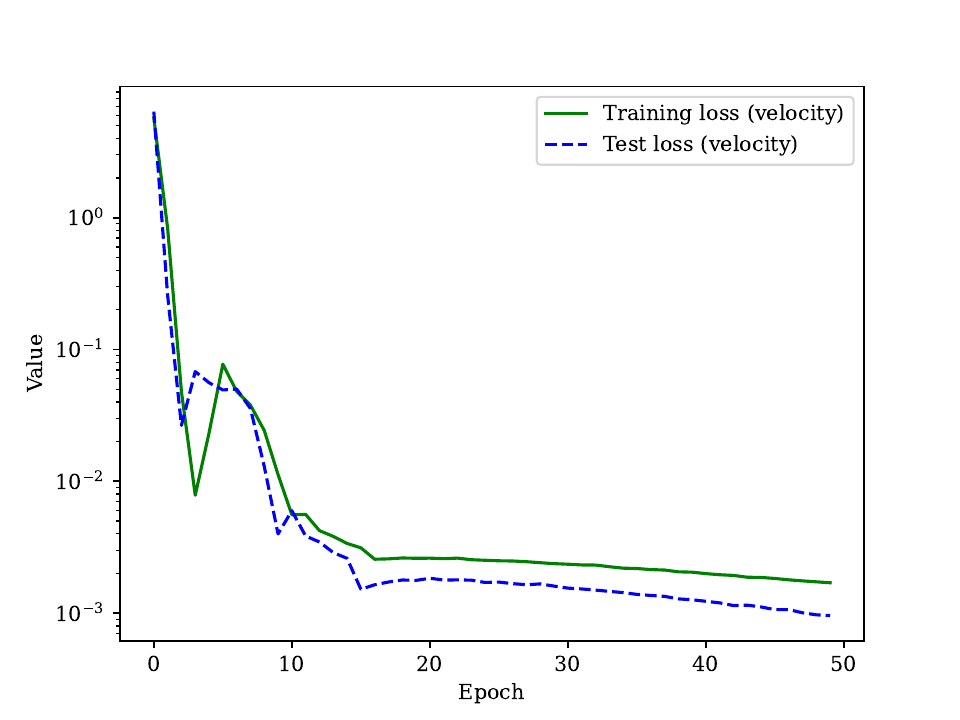}
    \includegraphics[width=0.49\linewidth]{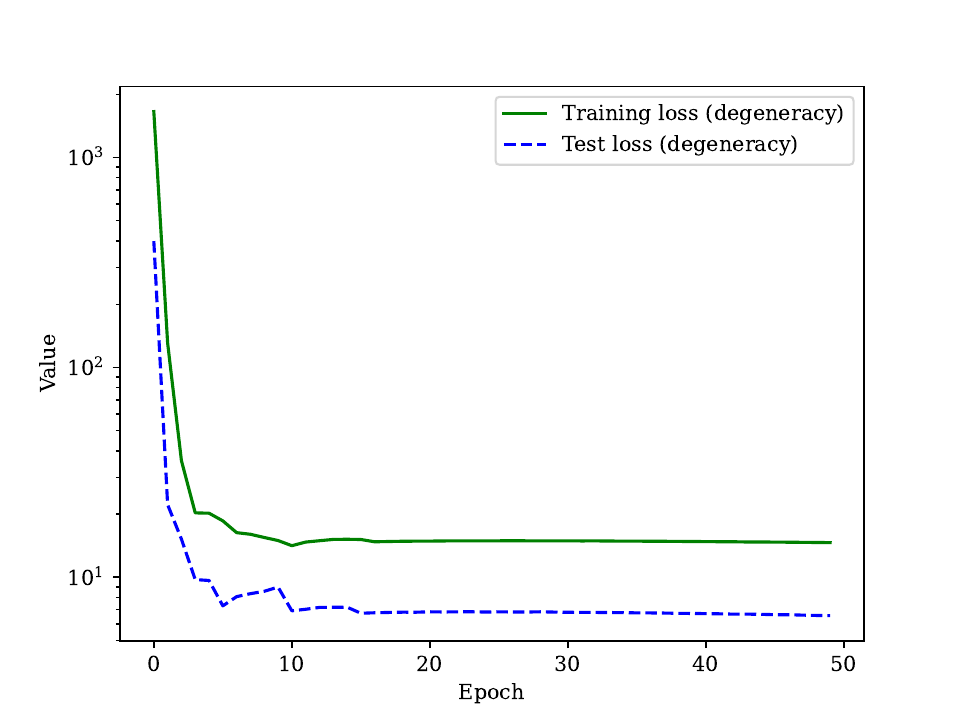}
    \caption{Evolution of loss functions for data (left) and degeneracy conditions (right) for the second artificial case.}
    \label{fig: loss case noise}
\end{figure}

\begin{figure}[htb]
    \centering
    \includegraphics[width=0.49\linewidth]{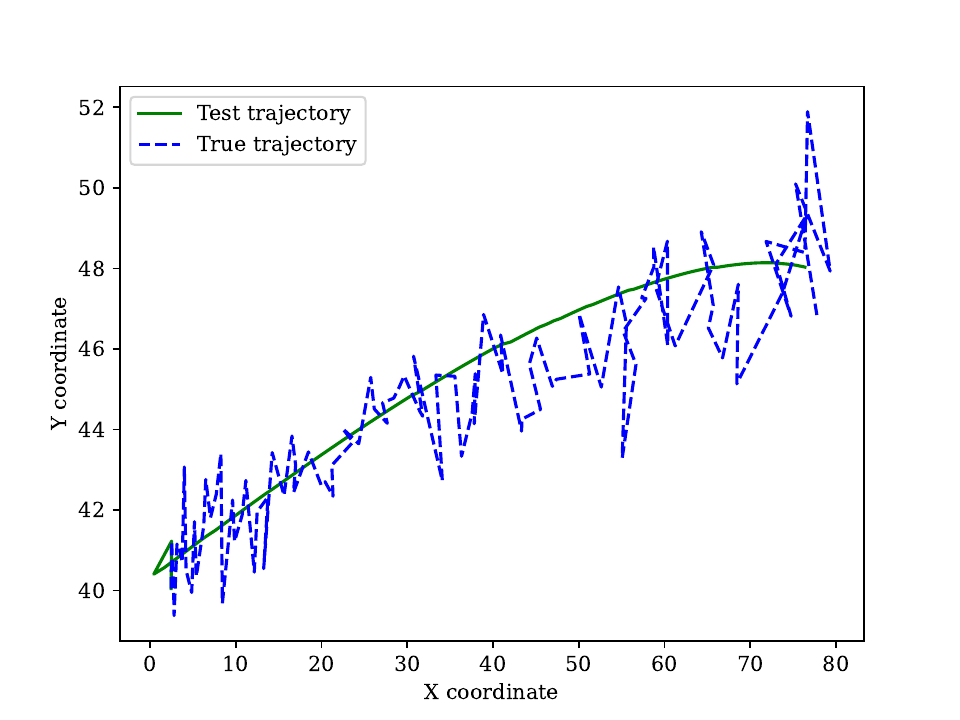}
    \includegraphics[width=0.49\linewidth]{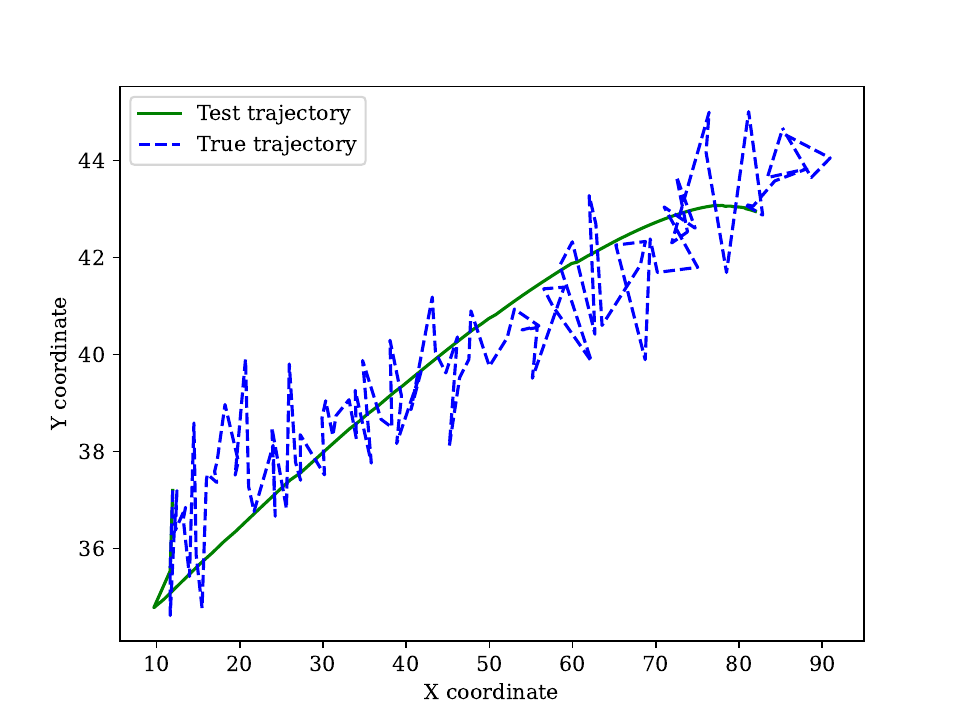}
    \caption{Comparison of two predicted and real trajectories for the second artificial case.}
    \label{fig: traj case noise}
\end{figure}

\begin{figure}[htb]
    \centering
    \includegraphics[width=0.55\linewidth]{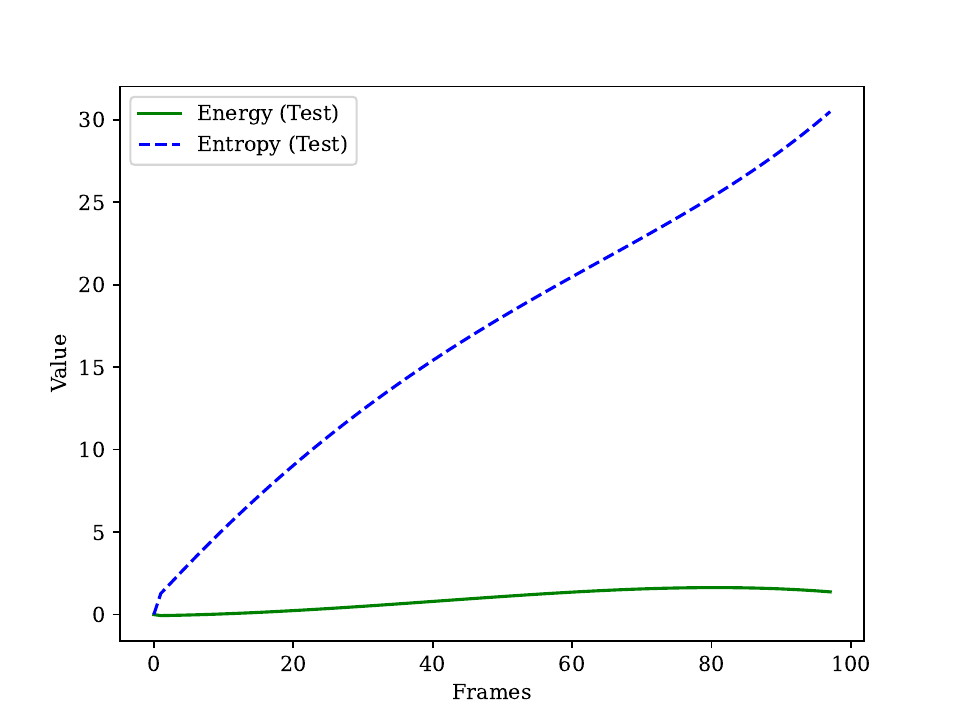}
    \caption{Evolution of thermodynamical variables for a complete trajectory from the second artificial case.}
    \label{fig: thrmd case noise}
\end{figure}

The results show that the model was still able to learn and reduce the loss function (see Figure \ref{fig: loss case noise}), with trajectories being correctly predicted (see Figure \ref{fig: traj case noise}) and thermodynamical variables evolving adequately (see Figure \ref{fig: thrmd case noise}).


The predicted trajectories do not simulate random variations in the noisy data but follow the same trend and capture the mean movement of the cell, which means that the model is able to correctly extract trajectories from inputs with noise. Even with the high levels of noise that can be appreciated in the true trajectory, the model is capable of filtering that noise and obtaining the mean behavior of the cell.

\begin{figure}[htb]
    \centering
    \includegraphics[width=0.49\linewidth]{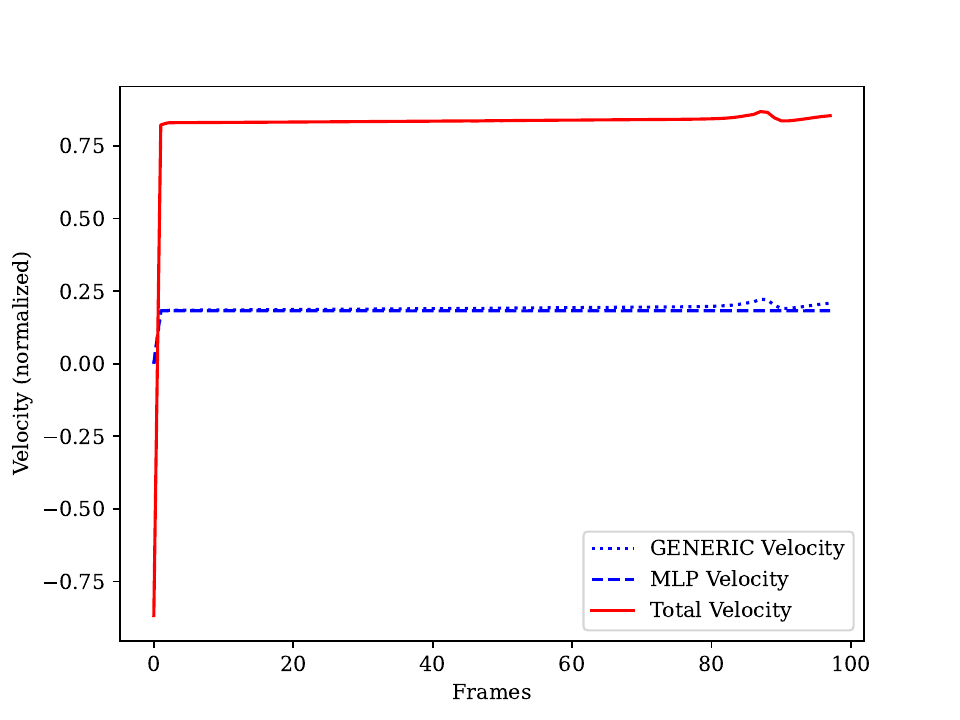}
    \includegraphics[width=0.49\linewidth]{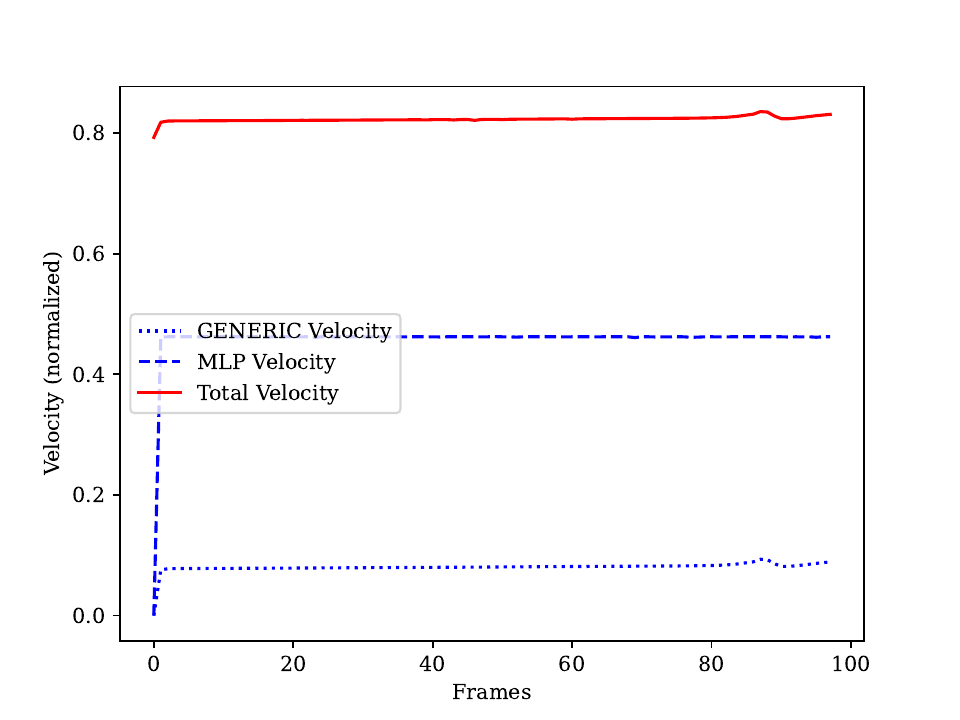}
    \caption{Contribution of each part of the model to the final output velocities for the \textit{in-silico} case with noise. SPNN shows an increasing tendency for both coordinates, while MLP provides variations.}
    \label{fig: vels case noise}
\end{figure}

Regarding the contribution of each part of the model to the final velocity output (see Figure \ref{fig: vels case noise}), a similar behavior to that of the previous case can be appreciated. In this case, in the $x$ axis, SPNN and MLP show a similar contribution, with SPNN capturing the final variations in the trajectory.

\FloatBarrier

\subsection{Real \textit{in vitro experiment}}
Once the performance of the model has been tested on simpler artificial cases, it can be adapted to the real experiment. As the real movement of cells is known to be highly dependent on environmental conditions, the initial weights of the MLP architecture are higher than in artificial cases. In addition, given the higher complexity and variability of real trajectories, the learning rates have been reduced to allow the model to undergo a slower but more confident learning process. Training parameters are shown in Table \ref{tab: params case real}.

\begin{longtable}{|c|c|c|c|}
\caption{Training parameters for the real experiment.}
\label{tab: params case real} \\
\hline
\centering
\textbf{Model} & \textbf{LR} & \textbf{LR Scheduler epochs} & \textbf{LR Scheduler $\gamma$} \\ \hline
\endhead
SPNN            & $5\cdot10^{-5}$ & $100$ & $0.9$ \\ \hline
MLP             & $5\cdot10^{-4}$ & $100$ & $0.9$ \\ \hline
Final MLP Layer & $5\cdot10^{-4}$ & $100$ & $0.9$ \\ \hline
\end{longtable}

\begin{figure}[htb]
    \centering
    \includegraphics[width=0.49\linewidth]{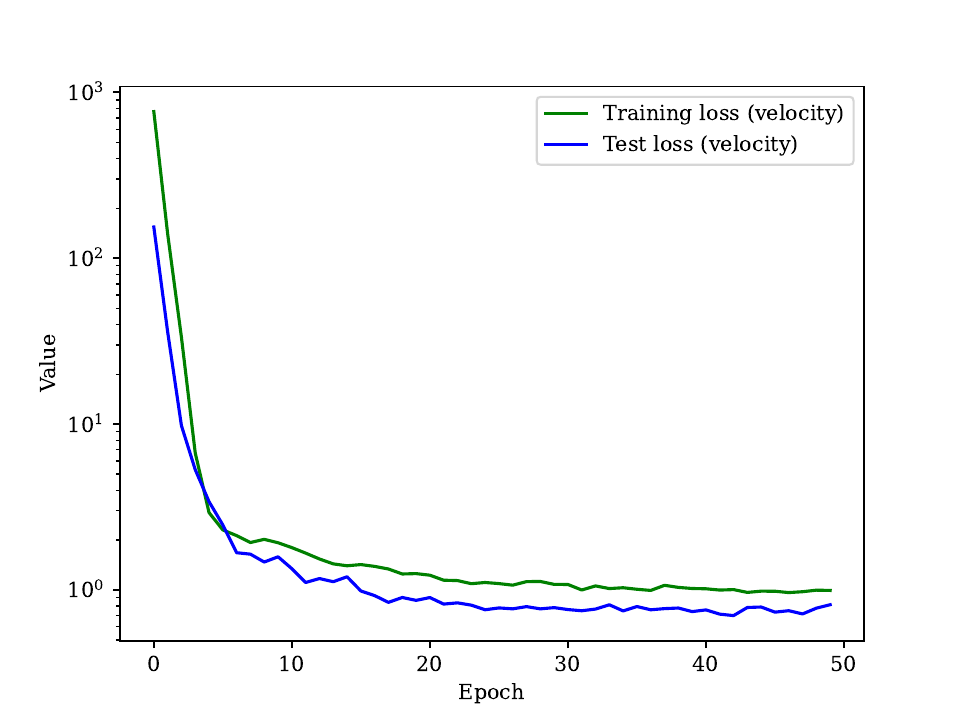}
    \includegraphics[width=0.49\linewidth]{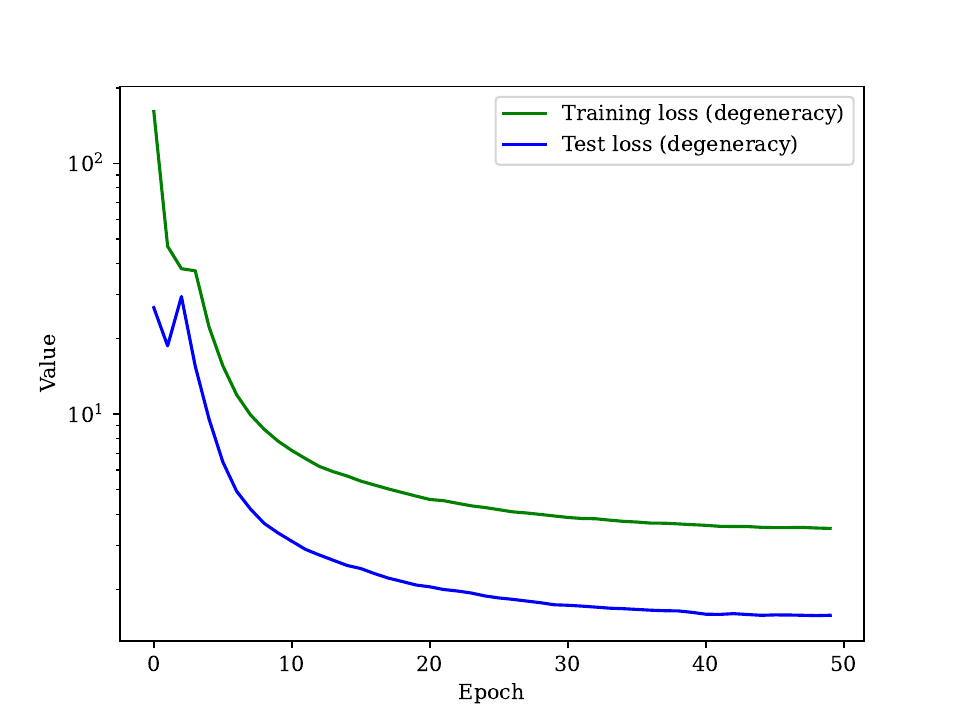}
    \caption{Evolution of loss functions for data (left) and degeneracy conditions (right) for the real experiment.}
    \label{fig: loss case real}
\end{figure}

\begin{figure}[htb]
    \centering
    \includegraphics[width=0.49\linewidth]{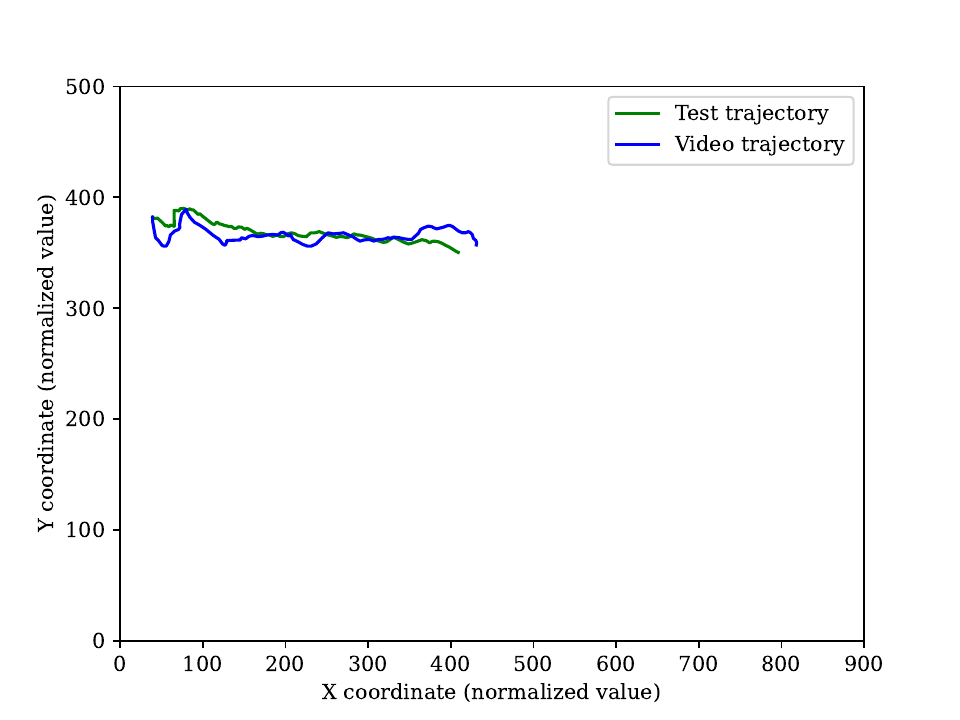}
    \includegraphics[width=0.49\linewidth]{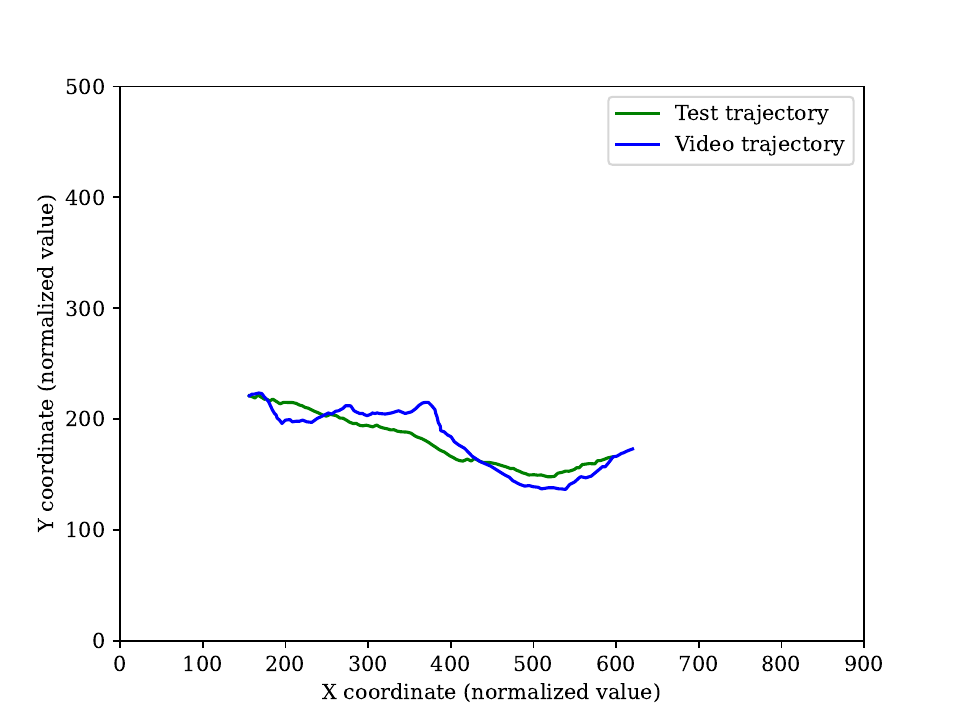}
    \includegraphics[width=0.49\linewidth]{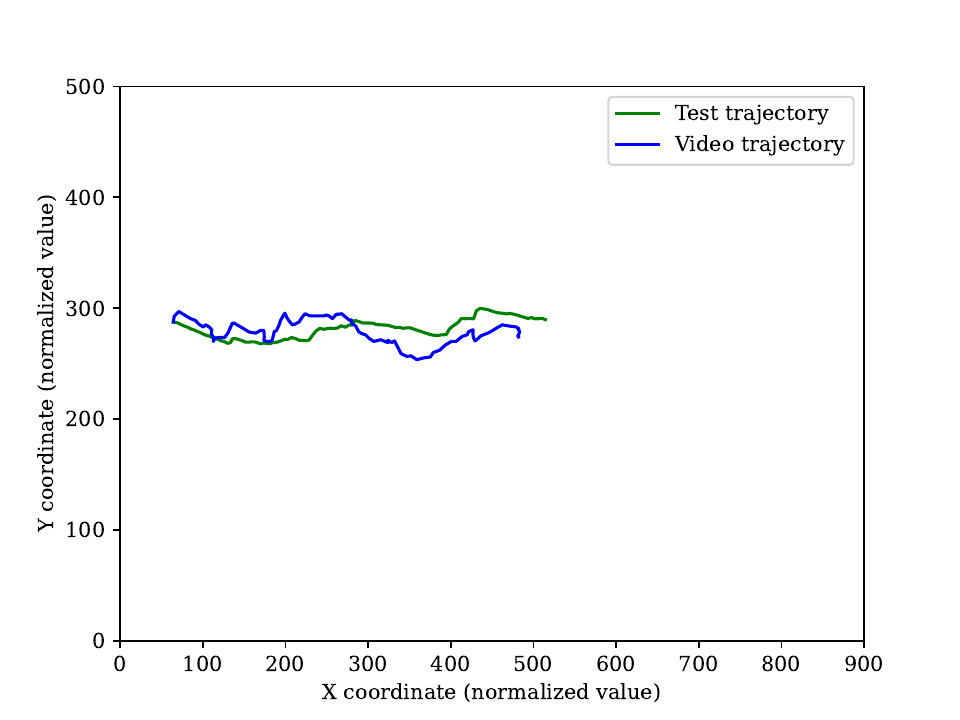}
    \includegraphics[width=0.49\linewidth]{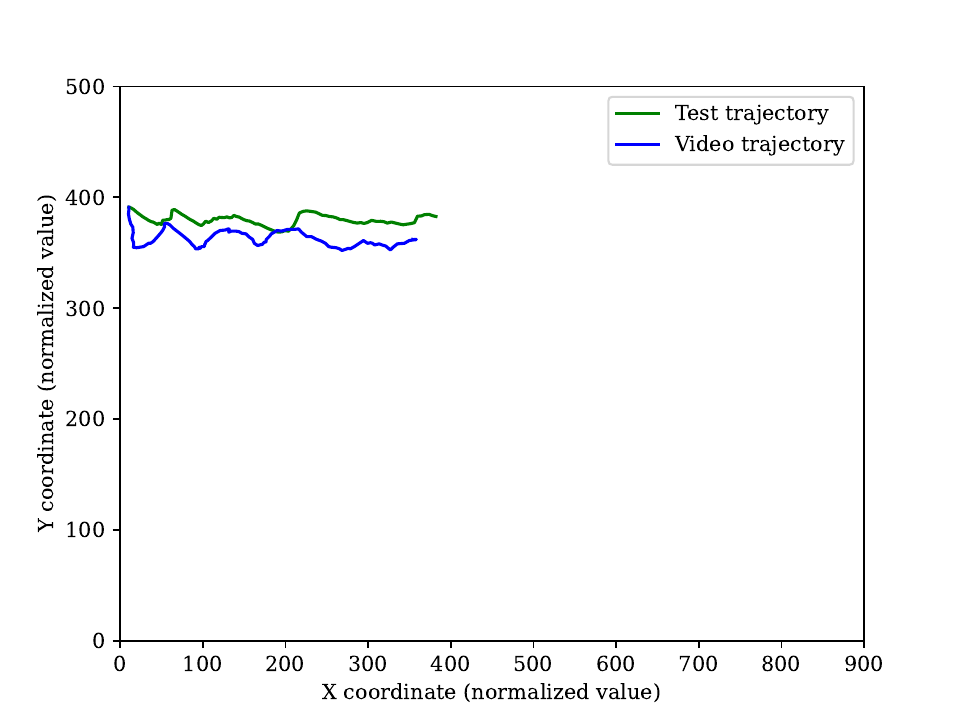}
    \caption{Comparison of four predicted and true trajectories for the real experiment.}
    \label{fig: traj case real}
\end{figure}

The model has shown very efficient learning capabilities, achieving accuracy levels of $91.2\%$ and $95.4\%$ for $x$ and $y$ coordinates, respectively. It is noticeable that the evolution of the loss functions (Figure \ref{fig: loss case real}) is much smoother than for the artificial cases and results in an optimized architecture. This shows the validity of reduced learning rates. Trajectories are being correctly predicted (see Figure \ref{fig: traj case real}) and accumulated error due to the \textit{roll-out policy} is not a concern, since final positions after $105$ frames are sufficiently close for real and predicted trajectories. In this case, the model is capable of representing velocity variations beyond simple two-dimensional gradients, correctly adapting to a fluctuating movement, especially in the $y$ coordinate, which shows more changes of direction than the $x$, coordinate, with a clear tendency (left-to-right).

\begin{figure}[htb]
    \centering
    \includegraphics[width=0.49\linewidth]{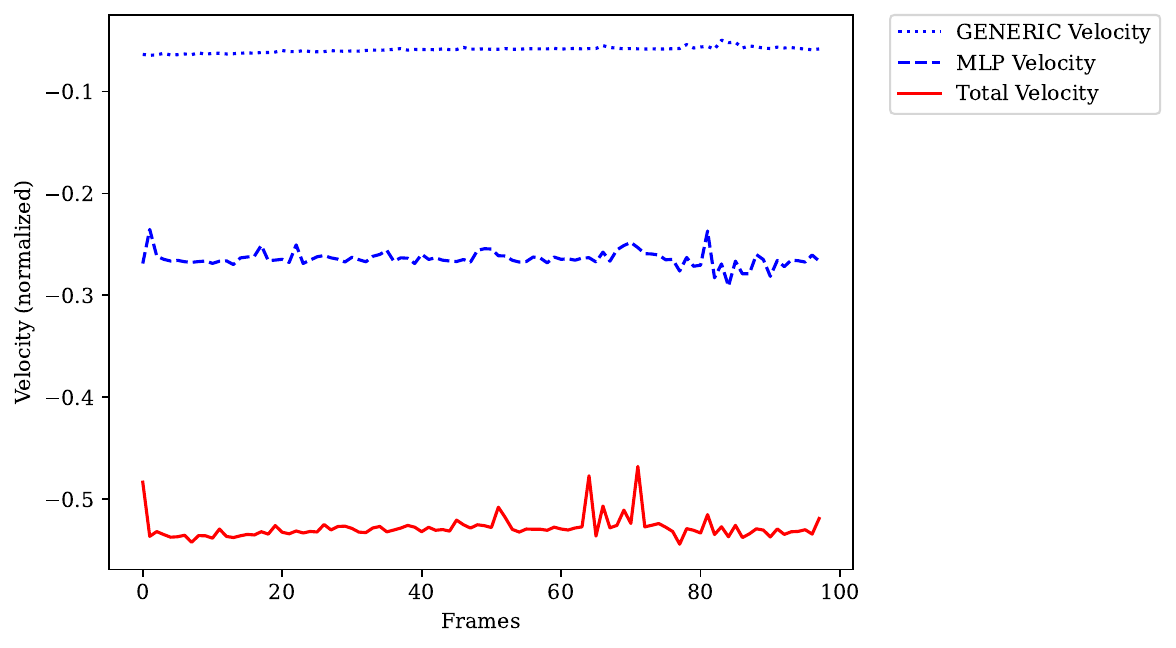}
    \includegraphics[width=0.49\linewidth]{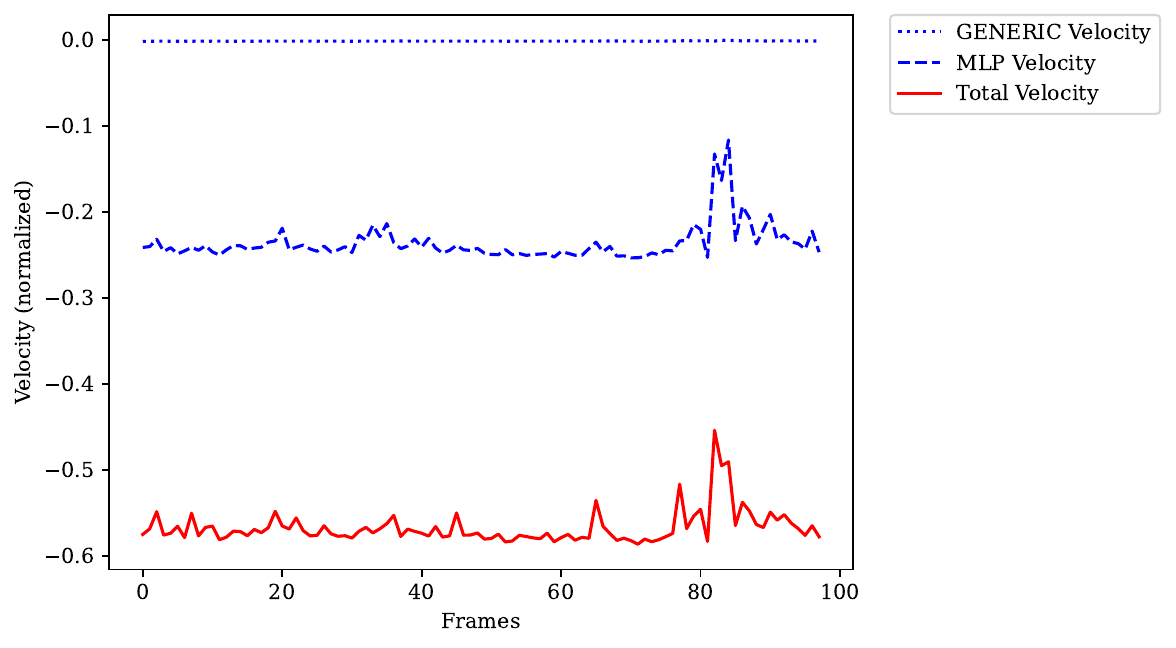}
    \caption{Contribution of each part of the model to the final output velocities for the real experiment.}
    \label{fig: vels case real}
\end{figure}

As stated above, real movements involve a much more complex mechanism. However, the clear trend observed in the $x$ coordinate could be related to the existence of a causing gradient (density gradient, since the right side of the microchannel is empty). The architecture of the complete model can be helpful in detecting which factors have the greatest influence on cell movement. As seen in Figure \ref{fig: vels case real}, velocities experience more variations over time (which aligns with real trajectories), and especially variations in MLP velocity are greater than in artificial cases. This reveals a higher importance of MLP input features (environmental conditions) in cell movement. It is noticeable that the MLP is dealing with heterogeneous variations in the trajectory, which could be caused by changing features of the environment (surrounding cells, surface density, internal energetic state of the cell, etc.), while the SPNN maintains a constant contribution that could be related to underlying gradients or movement causes.

In any case, this last analysis further confirms the main idea of this work, which is the need of an additional architecture that accounts for environmental features which are well beyond the Lagrangian behavior of the mechanical system. While the importance of this part of the model was not high, it captures most of the evolution of the system in the real \textit{in-vitro} case. The GENERIC architecture still shows a non-negligible contribution, which could be further related to constant environmental characteristics influencing cell movement.

\FloatBarrier

\begin{figure}[htb]
    \centering
    \includegraphics[width=0.6\linewidth]{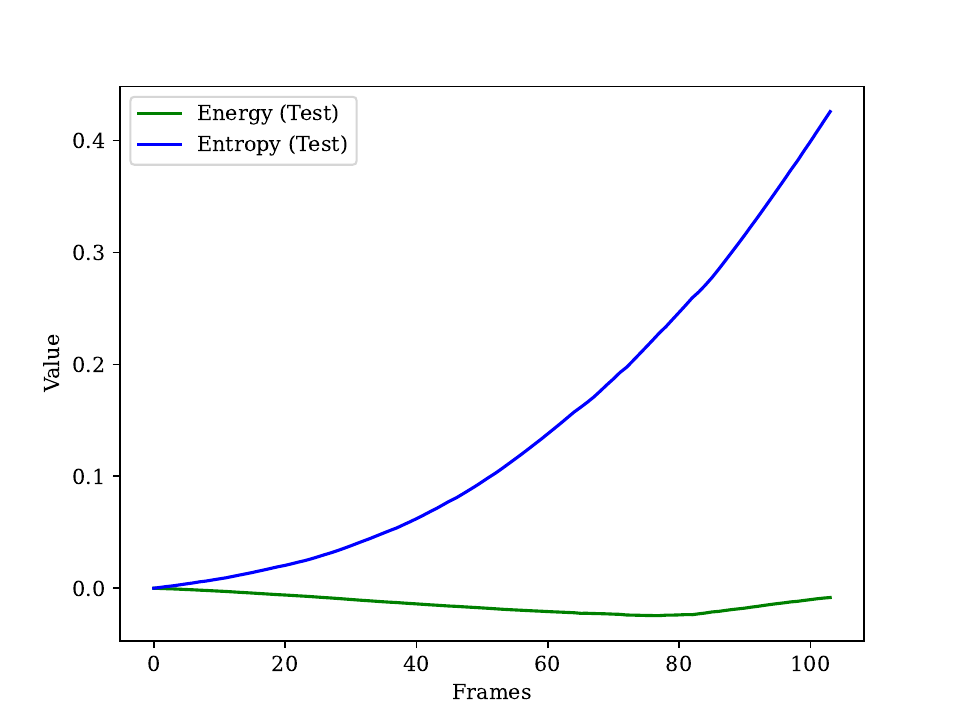}
    \caption{Evolution of thermodynamical variables for a complete trajectory from the real experiment.}
    \label{fig: thrmd case real}
\end{figure}

The model also represents the compliance of the system with thermodynamical principles, with a clear increase in entropy and a stable evolution (although a decreasing tendency can be observed) of total energy, as seen in Figure \ref{fig: thrmd case real}.

\FloatBarrier

\subsection{Mitosis prediction}

The mitosis prediction model was trained simultaneously with the ASPNN in the real experiment. As explained above, the loss was calculated with BCE function, and the training parameters are shown in Table \ref{tab: params mitosis}. This model has been trained for $20000$ epochs, since preliminary tests showed that $5000$ epochs were not sufficient.

\begin{longtable}{|c|c|c|c|}
\caption{Training parameters for the mitosis prediction model.}
\label{tab: params mitosis} \\
\hline
\centering
\textbf{Model} & \textbf{LR} & \textbf{LR Scheduler epochs} & \textbf{LR Scheduler $\gamma$} \\ \hline
\endhead
Mitosis            & $2\cdot10^{-4}$ & $40000$ & $0.5$ \\ \hline
\end{longtable}

As stated before, variations in area and brightness have been considered fundamental inputs for this model. As mitosis processes have an extended duration over several frames, these variations were calculated for frames periods recursively, which means that variations between frames were added for a given frame. Thus, two new parameters were included in the model: the number of previous frames to consider when calculating the area and brightness recurrence. They were set to 2 and 3, recursively, as it was observed that increases in cell brightness began before variations in shape (and area), which is consistent with the description of mitosis processes in the literature \cite{georgi2002timing}.

\begin{figure}[htb]
    \centering
    \includegraphics[width=0.6\linewidth]{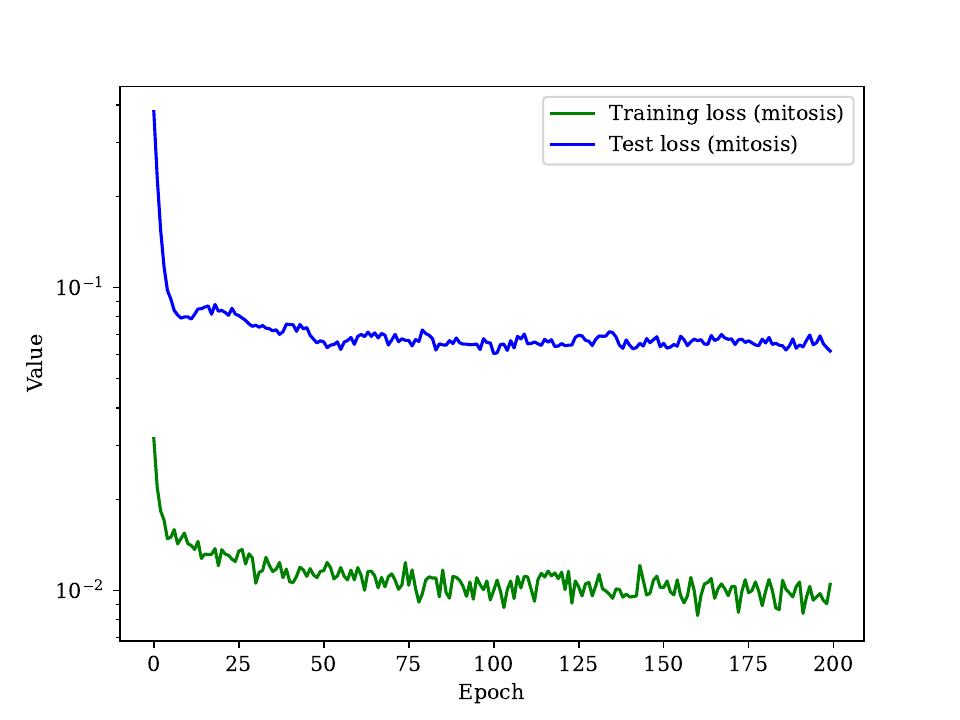}
    \caption{Evolution of loss function for mitosis prediction model.}
    \label{fig: loss mitosis}
\end{figure}

The resulting model has been tested on a set of $30$ trajectories with a total of $12$ mitotic events. The model was able to correctly learn and reduce the loss function (see Figure \ref{fig: loss mitosis}).

\begin{figure}[htb]
    \centering
    \includegraphics[width=0.49\linewidth]{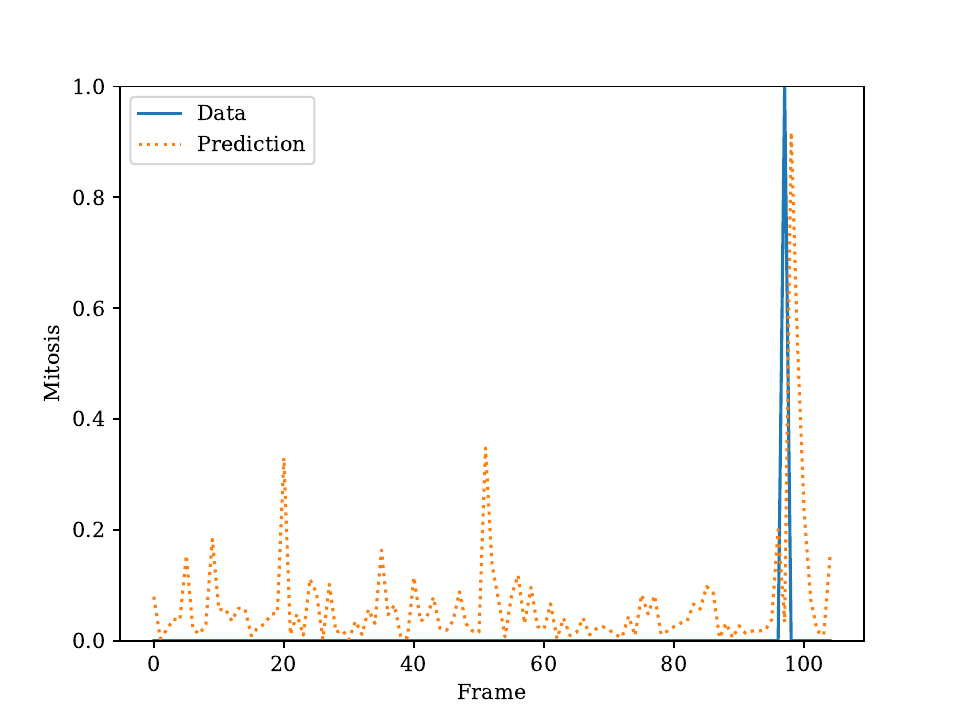}
    \includegraphics[width=0.49\linewidth]{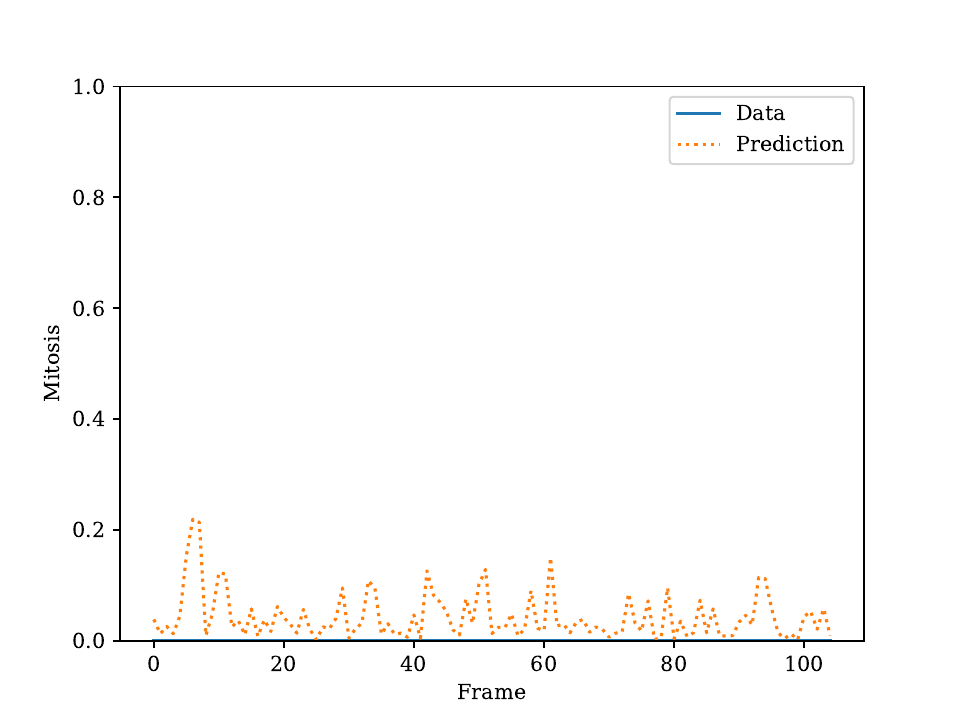}
    \includegraphics[width=0.49\linewidth]{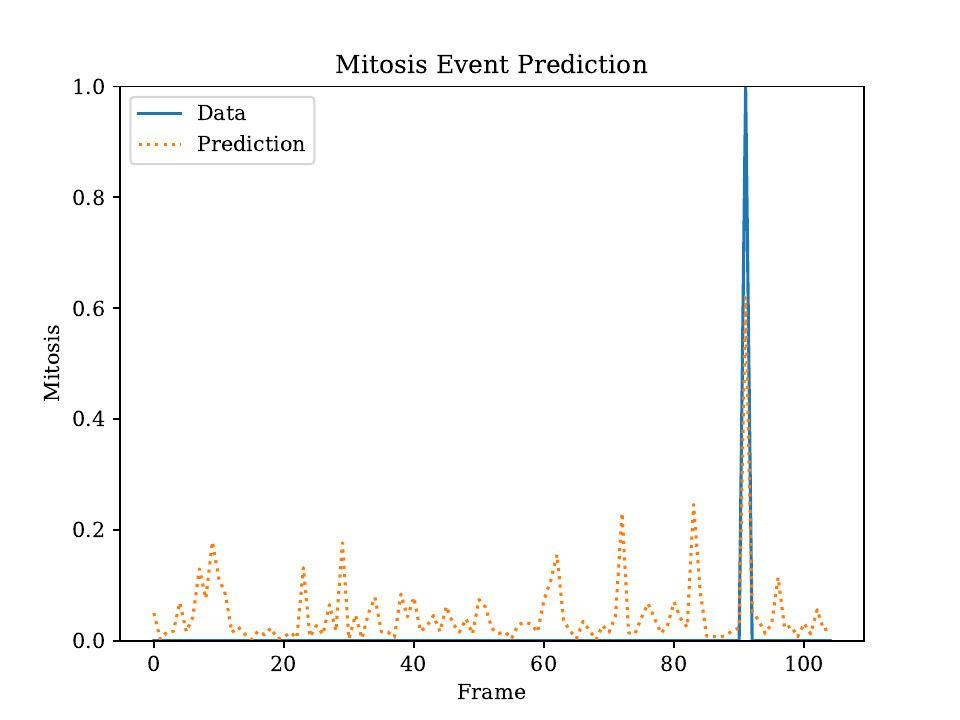}
    \includegraphics[width=0.49\linewidth]{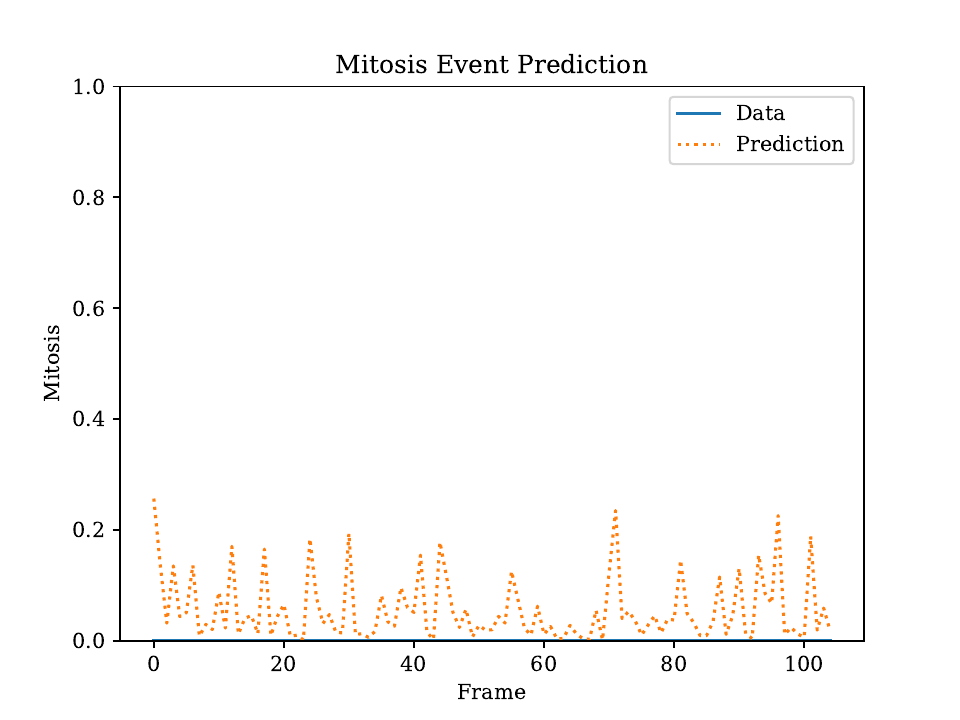}
    \caption{Comparison of model outputs and ground truth data for two trajectories with one mitotic event (left) and two without any mitotic events (right).}
    \label{fig: mitosis}
\end{figure}

The model achieved a precision of $69\%$, defined as correctly predicted positive events out of total positive events, in the test set. An event is correctly predicted if the probability of mitosis in a $6$ frame window ($3$ previous and $3$ next frames) centered in the frame with a real mitosis is higher than $0.6$ (within the $[0,1]$ interval). The percentage of false positive predictions was $0.1\%$, which shows that the model has adequate generalization capabilities and does not trigger easily. Two examples of comparison of model outputs with ground truth data are shown in Figure \ref{fig: mitosis}. 

\FloatBarrier

\section{Conclusions}

The presented work provides a new approach to the analysis of cell migration. It proposes a novel framework that recognizes the complexity of biological systems, and specifically cell networks and clusters, which move according to a large number of mechanisms that involve their own state and the environmental conditions of their surroundings. We propose a new tool for studying many of these factors together, which eliminates the need for complex experiment designs aimed at highlighting a particular feature.

Our Augmented Structural Preserving Neural Network is based on two different parts: an SPNN and an MLP architecture. These two parts each focus on a part of the biological problem. While the SPNN studies the movement of cells as purely mechanical objects which must comply with thermodynamical principles, the MLP accounts for existing patterns in cell behavior that might be a response to its surroundings. The first one only takes positional and velocity parameters as inputs, while the second one uses a set of environmental characteristics that can be extracted from the visual analysis of cell movement in a video. In order to obtain these data, Computer Vision techniques involving image segmentation (SAM model) for cell detection and object tracking for trajectory determination (DeepOCSort) have been used to obtain cell positions over time, which have later been processed to obtain environmental conditions for each cell in each frame. The combined use of these two architectures with different inputs can also provide insight into the influence of each set of inputs on cell movement.

The model has been tested on three cases with an increasing degree of complexity: an artificial case without noise, an artificial case with noise, and a real \textit{in vitro} experiment. For all three cases, velocity accuracies have exceeded $85\%$ in both coordinates, with higher values in the first artificial case and the real experiment. Analysis of trajectories has shown that the accumulated error due to the \textit{roll-out} policy is not a major concern. Moreover, compliance with thermodynamical principles has also been demonstrated.

Lastly, we present a mitosis event prediction model based on an MLP architecture that uses the same visually observed features as the ASPNN as inputs. This model has achieved $69\%$ accuracy, which reveals that mitosis prediction is possible. Future lines of work should more deeply analyze this possibility, as being able to foresee cell division and proliferation mechanics could represent a major advance in the understanding of many biological processes or pathologies.
\label{sec:conclusion}

\section{Acknowledgements}
Grant PID2021-126051OB-C43 funded by MCIN/AEI/ 10.13039/ 501100011033 and “ERDF A way of making Europe”.

 \bibliographystyle{elsarticle-num} 
 \bibliography{references}
\end{document}